\documentclass[10pt,twocolumn,letterpaper]{article}

\usepackage{cvpr}
\usepackage{times}
\usepackage{epsfig}
\usepackage{graphicx}
\usepackage{amsmath}
\usepackage{amssymb}
\usepackage{subfigure}
\usepackage{multirow}

\usepackage[pagebackref=true,breaklinks=true,letterpaper=true,colorlinks,bookmarks=false]{hyperref}

\cvprfinalcopy 

\def\cvprPaperID{2374} 

\ifcvprfinal\fi
\begin{document}

\title{Multi-Scale Geometric Consistency Guided Multi-View Stereo}

\author{Qingshan Xu and Wenbing Tao\thanks{Corresponding author}\\
National Key Laboratory of Science and Technology on Multispectral Information Processing\\
School of Artifical Intelligence and Automation, Huazhong University of Science and Technology, China\\
{\tt\small \{qingshanxu, wenbingtao\}@hust.edu.cn}
}

\maketitle
\thispagestyle{empty}

\begin{abstract}
   In this paper, we propose an efficient multi-scale geometric consistency guided multi-view stereo method for accurate and complete depth map estimation. We first present our basic multi-view stereo method with Adaptive Checkerboard sampling and Multi-Hypothesis joint view selection (ACMH). It leverages structured region information to sample better candidate hypotheses for propagation and infer the aggregation view subset at each pixel. For the depth estimation of low-textured areas, we further propose to combine ACMH with multi-scale geometric consistency guidance (ACMM) to obtain the reliable depth estimates for low-textured areas at coarser scales and guarantee that they can be propagated to finer scales. To correct the erroneous estimates propagated from the coarser scales, we present a novel detail restorer. Experiments on extensive datasets show our method achieves state-of-the-art performance, recovering the depth estimation not only in low-textured areas but also in details. 
\end{abstract}

\section{Introduction}\label{Sec:Int}

Multi-view stereo (MVS) has traditionally been a topic of interest in computer vision
for decades. It aims at establishing dense correspondence from multiple calibrated images, which results in a dense 3D reconstruction. Over the last few years, much effort has been put into improving the quality of dense 3D reconstructions and some works have achieved impressive results~\cite{Furukawa2010Accurate,Galliani2015Massively,Goesele2007Multi,Shan2013Visual,Shan2014Occluding,Shen2013Accurate,Zheng2014PatchMatch,Schonberger2016Pixelwise}. However, with the large-scale data, low texture, occlusions, repetitive patterns and reflective surface, it is still a challenging problem to perform efficient and accurate multi-view stereo in computer vision domain.   

\begin{figure}[t]
	\setlength{\abovecaptionskip}{0pt}
	\setlength{\belowcaptionskip}{0pt}
	\centering
	\subfigure[]{
		\label{fig:subfig:a}
		\includegraphics[width=0.315\linewidth]{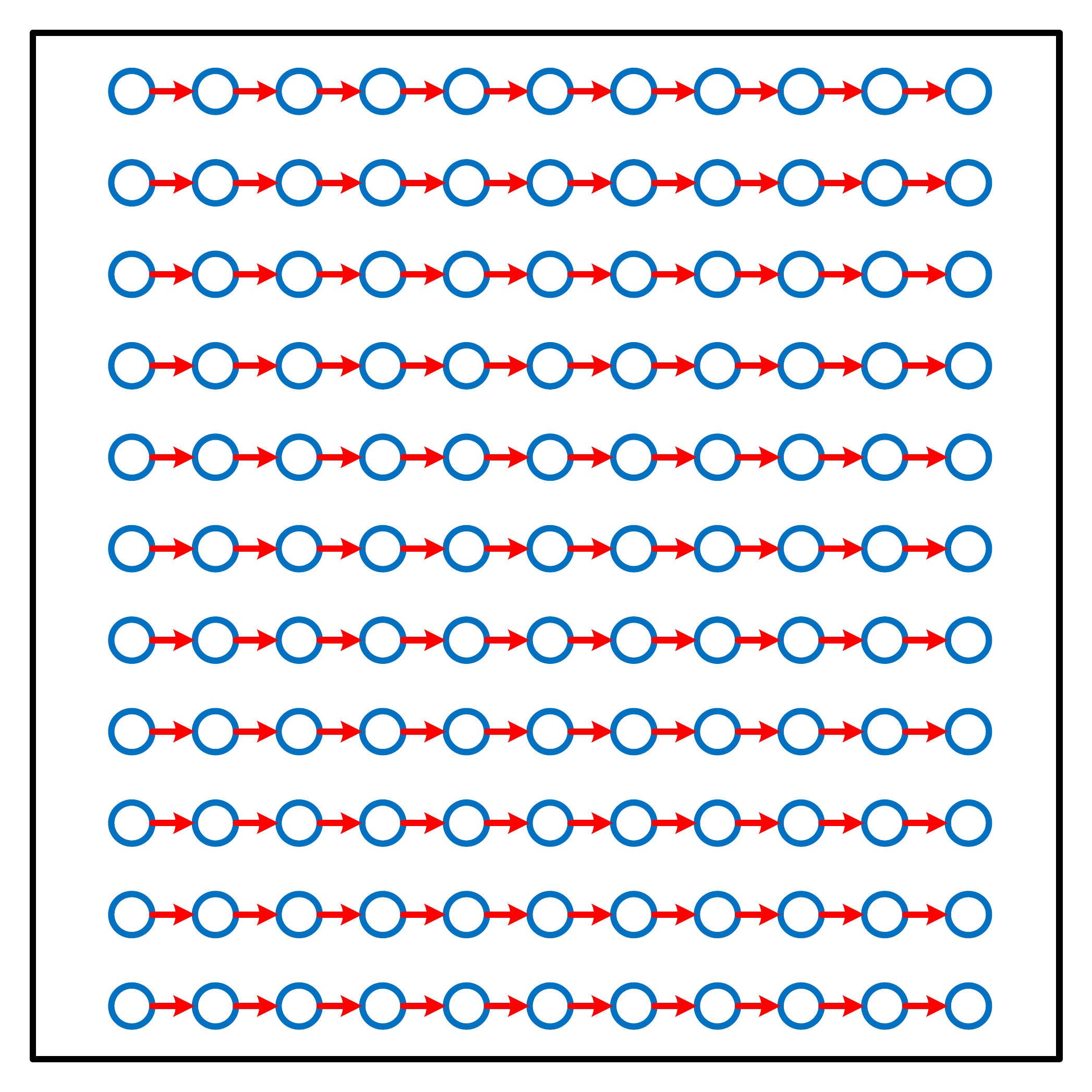}}
	\subfigure[]{
		\label{fig:subfig:b}
		\includegraphics[width=0.315\linewidth]{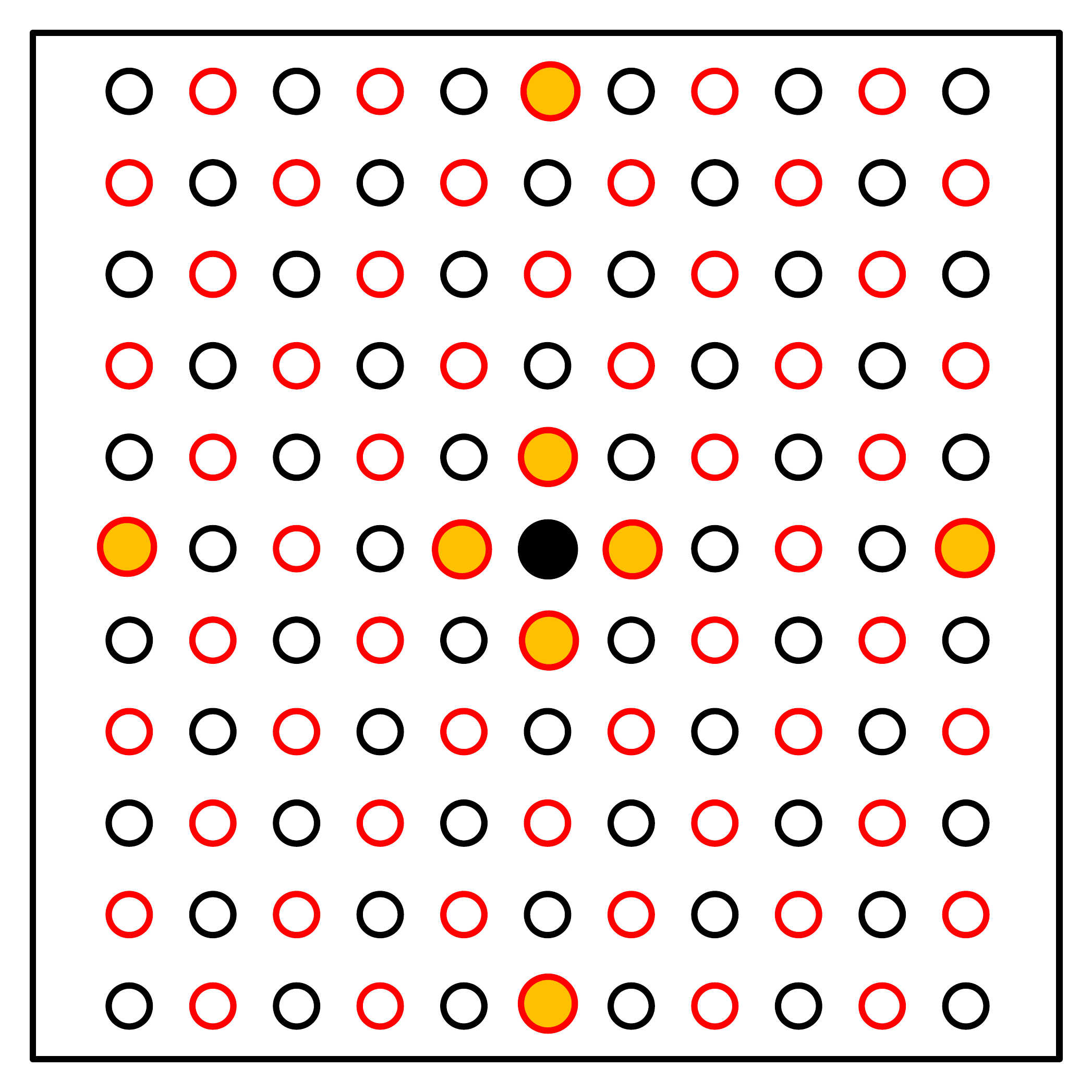}}
	\subfigure[]{
		\label{fig:subfig:c}
		\includegraphics[width=0.315\linewidth]{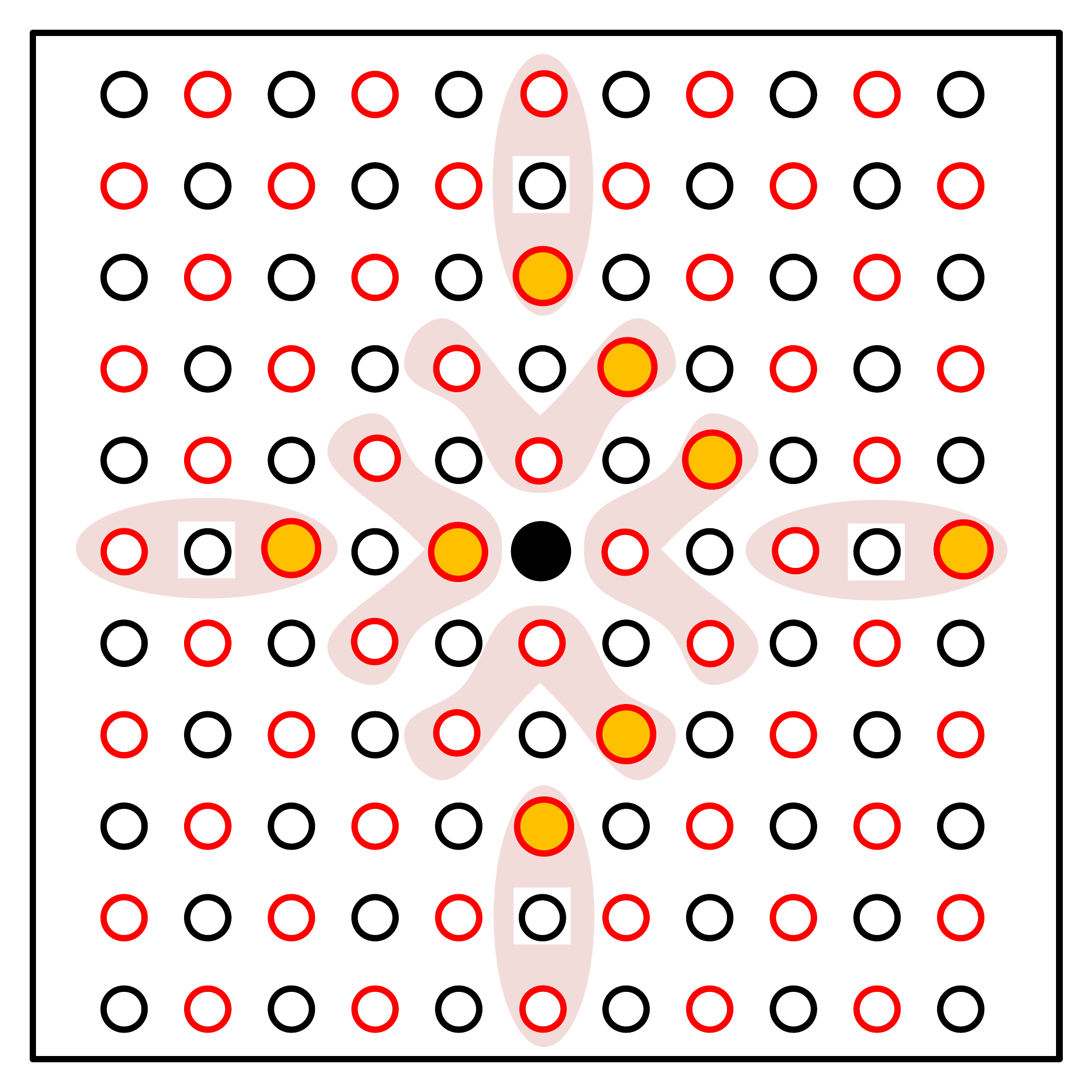}}
	\caption{Propagation scheme. (a) Sequential propagation. (b) Symmetric checkerboard propagation. (c) Adaptive checkerboard propagation. The light red areas in (c) show sampling regions. The solid yellow circles in (b) and (c) show the sampled points.}
	\label{fig:ACMM}
\end{figure} 

Recently, PatchMatch Stereo methods~\cite{Bailer2012Scale,Zheng2014PatchMatch,Galliani2015Massively,Schonberger2016Pixelwise} show great power in depth map estimation with their fast global search for the best match in other images~\cite{Barnes2009PatchMatch}. These methods follow a popular four-step pipeline, including random initialization, propagation, view selection and refinement. In this pipeline, propagation and  view selection are two key steps to PatchMatch Stereo methods. The former is important to efficiency while the latter is critical to accuracy. 

For propagation, there generally exist two distinct types of parallel schemes: sequential propagation~\cite{Bailer2012Scale,Zheng2014PatchMatch,Schonberger2016Pixelwise} and diffusion-like propagation~\cite{Galliani2015Massively}. The former traverses pixels following parallel scanlines only in the vertical (or horizontal) direction (Figure~\ref{fig:subfig:a}). In contrast, the later simultaneously updates the status of half of the pixels in an image with a checkerboard pattern (Figure~\ref{fig:subfig:b}). In terms of efficiency, the diffusion-like propagation achieves better algorithm parallelism. However, it is reported in \cite{Schonberger2016Pixelwise,Schops2017Multi} that, its reconstruction results are not competitive with the sequential propagation's in some challenging cases. 
As pointed out in \cite{Zheng2014PatchMatch}, this mainly attributes to its less robust view selection instead of propagation. For example, in the sequential propagation, \cite{Zheng2014PatchMatch,Schonberger2016Pixelwise} construct a probabilistic graphical model to perform pixelwise view selection. Unlike their elaborate view selection, the diffusion-like propagation adopts a simple threshold truncation scheme to determine aggregation view subsets~\cite{Galliani2015Massively}. This leads to its biased view selection for different hypotheses. Then a motivating question is, whether it is possible to design a more robust view selection based on the checkerboard pattern.

To this end, we first propose our basic MVS method with Adaptive Checkerboard sampling and Multi-Hypothesis joint view selection (ACMH). Our key idea is based on the assumption of \cite{Bleyer2011PatchMatch} that pixels within a relatively large region can be approximately modeled by one 3D plane, which indicates \emph{structured region information} and a
shared hypothesis among these pixels. Thus, unlike fixed sampling in diffusion-based conventions which may be misleading, ACMH searches larger regions to adaptively sample better candidate hypotheses for propagation (Figure~\ref{fig:subfig:c}). With these better hypotheses, we propose a multi-hypothesis joint strategy to infer pixelwise view selection. For a specific pixel, this strategy employs a voting scheme to supply the same aggregation view subset for different propagated hypotheses, and gives credible views greater weights to aggregate the final multi-view matching cost. As a result, ACMH can achieve accurate depth map estimation while inheriting the high efficiency of the checkerboard pattern.

\begin{figure}[t]
	\setlength{\abovecaptionskip}{0pt}
	\setlength{\belowcaptionskip}{0pt}
	\centering
	\includegraphics[width=0.99\linewidth]{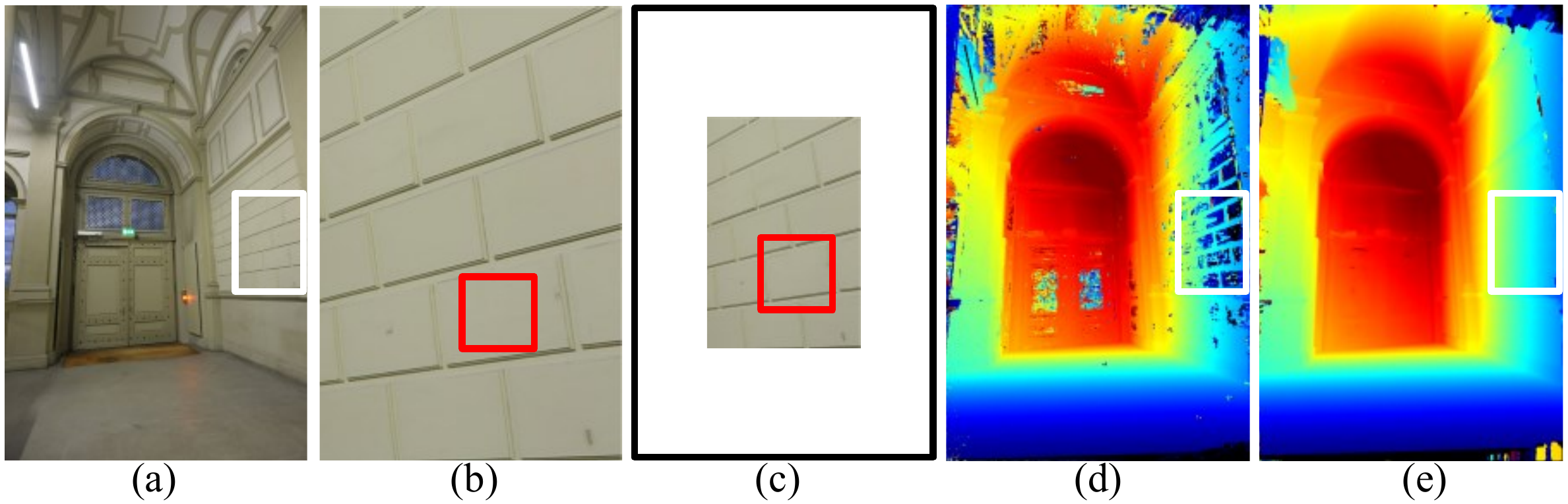}
	\caption{Texture richness for different scales. (a) Original Image. (b) The zoomed-in version of the white box in (a). (c) The downsampled version of (b). (d) Depth map obtained with the original scale. (e) Depth map obtained with the multi-scale scheme. The patch windows in red are kept the same size.}
	\label{fig:MS}
\end{figure}

Moreover, as a key component of PatchMatch Stereo methods, view selection heavily depends on a stable visual similarity measure between two image patches. However, measuring the visual similarity in low-textured areas is always challenging. As depicted in Figure~\ref{fig:MS}(b), the low discrimination in low-textured areas leads to the ambiguity of visual similarity, which further degrades the performance of PatchMatch Stereo methods (Figure~\ref{fig:MS}(d)). However, we observe that, for the low-textured areas, though the texture information with an universal patch window in Figure~\ref{fig:MS}(b) is not significant , it becomes more discriminative under the same patch window when an image is downsampled (Figure~\ref{fig:MS}(c)). That is, the texture richness is a relative measure. Then, an intuitive idea is that, we can estimate depth information at coarser scales to alleviate the ambiguities in low-textured areas and use it as guidance for the matching progress at finer scales.        

Based on the above idea, we further present a multi-scale patch matching with geometric consistency guidance, called ACMM. Specifically, our method constructs image pyramids and obtains reliable depth estimates for low-textured areas at coarser scales. After propagating these estimates from coarser scales to finer scales via upsampling, we resort to geometric consistency to constrain the depth optimization at finer scales. Considering that the depth propagation from coarser scales to finer scales often leads to depth information loss in details, we present a detail restorer based on the difference map of photometric consistency between adjacent scales. Through our proposed strategies, our approach can not only estimate depth information in low-textured areas but also preserve details.  

Our main contributions are summarized as follows: 1) Inherited from the high efficiency of the diffusion-like propagation, we present an adaptive checkerboard sampling scheme to select more reasonable hypotheses for propagation based on the structured region information. Then, a multi-hypothesis joint view selection is proposed to help select credible aggregation views. 2) For the ambiguities in low-textured areas, we propose a multi-scale patch matching scheme with geometric consistency guidance. The geometric consistency imposed at different scales can guarantee that the reliable depth estimates for low-textured areas obtained at coarser scales are retained at finer scales. Moreover, a detail restorer is present to correct errors propagated from the coarser scales. Through extensive evaluation, we demonstrate the effectiveness and efficiency of our method by achieving state-of-the-art performance on Strecha dataset~\cite{Strecha2008On} and ETH3D benchmark\cite{Schops2017Multi}.

\section{Related Work} 

\begin{figure*}[t]
	\setlength{\abovecaptionskip}{0pt}
	\setlength{\belowcaptionskip}{0pt}
	\centering
	\includegraphics[width=0.85\linewidth]{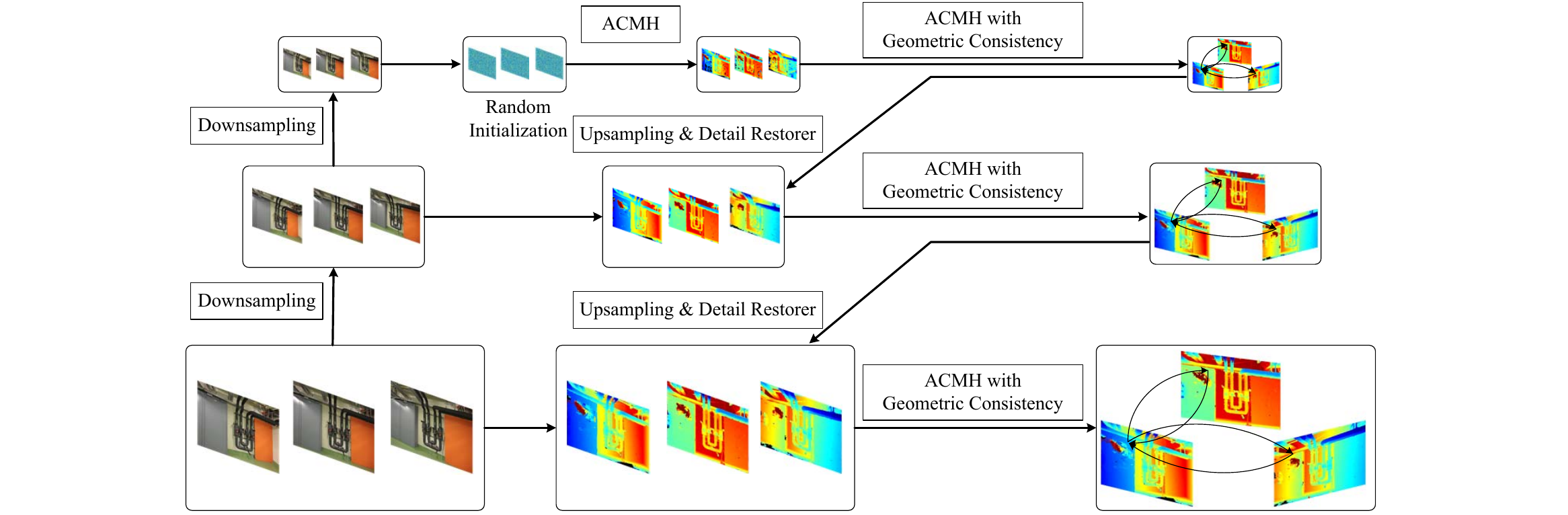}
	\caption{Overview of our approach. The initial depth maps of the coarsest scale are obtained by our basic MVS model with only photometric consistency (Section~\ref{Sec:ACMH}). After upsampling the estimation of the previous scale to the current scale, detail restorer is implemented to correct the errors in details. At each scale, geometric consistency is enforced to enhance coherence and prevent the reliable estimates in low-textured areas from the previous scale being impaired by photometric consistency (Section~\ref{Sec:PGC}).}
	\label{fig:Overview}
\end{figure*}

According to \cite{Seitz2006Comparison}, MVS methods can be categorized into four groups, voxel-based methods~\cite{Faugeras1998Variational,Vogiatzis2007Multiview,Sinha2007Multi}, surface evolution based methods~\cite{Hernandez2004Silhouette,Hiep2009Towards,Cremers2011Multiview}, patch-based methods~\cite{Goesele2007Multi,Lhuillier2005Quasi,Furukawa2010Accurate} and depth map based methods~\cite{Zheng2014PatchMatch,Galliani2015Massively,Schonberger2016Pixelwise}. The voxel-based methods are often constrained by their predefined voxel grid resolution. The surface evolution based methods depend on a good initial solution. As for the patch-based methods, its dependence on matched keypoints impairs the completeness of 3D models. The depth map based methods require estimating depth maps for all images and then fusing them into a unified 3D scene representation. A more detailed overview of MVS methods is presented in \cite{Seitz2006Comparison,Furukawa2015MST}. Our method belongs to the last category and we only discuss the related PatchMatch Stereo approaches. 

In terms of efficiency, \cite{Bailer2012Scale,Wei2014Multi,Zheng2014PatchMatch,Schonberger2016Pixelwise} adopt the sequential propagation scheme. They alternatively perform upward/downward propagation in odd iteration steps and perform leftward/rightward propagation in even steps. To increase parallelism, \cite{Wei2014Multi} selects an eighth of the image height (width) as the length of each scanline in the vertical (horizontal) propagation. However, the algorithm parallelism of sequential propagation is still proportional to the number of rows or columns of images. Then, Galliani \etal~\cite{Galliani2015Massively} propose to leverage a  checkerboard pattern to perform a diffusion-like propagation scheme. It allows to simultaneously update the status of half of the pixels in an image. However, they ignore good hypotheses should have priority in propagation.

According to the above propagation strategies, many view selection schemes are proposed to tackle the noise in the propagation process. In the diffusion-like propagation scheme, \cite{Galliani2015Massively} selects fixed $k$ views with the minimal $k$ matching costs. However, this leads to a bias due to different aggregation subsets for different hypotheses. In the sequential propagation, \cite{Bailer2012Scale,Wei2014Multi} also ignore the pixelwise view selection by only demanding global view angles. To incorporate only useful neighboring views at each pixel,  Zheng \etal~\cite{Zheng2014PatchMatch} first try to construct a probabilistic graphical model to jointly estimate depth maps and view selection. Further, Sch{\"o}nberger \etal~\cite{Schonberger2016Pixelwise} introduce geometric priors and temporal smoothness to better depict the state-transition probability. However, this sequential inference needs to condition the status of previous pixels at the current state. It is still more sensitive to noise in low-textured areas.  

Although some methods focus on view selection to improve local smoothness and gain some benefits, they are still restricted by patch window size. To perceive more useful information in low-textured areas, Wei \etal~\cite{Wei2014Multi} adopt the multi-scale patch matching with variance based consistency. However, this consistency is too strong to spread some reliable estimates in few neighboring views across multiple views. Moreover, they overlook the errors in details.

\section{Overview}
 
Given a set of input images $\mathcal{I}=\{\emph{I}_{i}\,|\,i=1{\cdots}N\}$ with known calibrated camera parameters $\mathcal{P}=\{\emph{P}_{i}\,|\,i=1{\cdots}N\}$, our goal is to estimate depth maps $\mathcal{D}=\{\emph{D}_{i}\,|\,i=1{\cdots}N\}$ for all images and fuse them into a 3D point cloud. Specifically, we aim to recover the depth map for reference image $I_{\text{ref}}$ sequentially selected from $\mathcal{I}$ with the guidance of source images $I_{\text{src}}$ ($\mathcal{I}-I_{\text{ref}}$). 

An overview of our method is illustrated in Figure~\ref{fig:Overview}. We construct a pyramid with $k$ scales for all images with a downsampling factor $\eta$. We denote the $l$-th scale of $\emph{I}_{i}$ and corresponding camera parameter as $\emph{I}_{i}^{l}$ and $\emph{P}_{i}^{l}$, $l=0{\cdots}k-1$. The finest scale of the pyramids $\emph{I}_{i}^{k-1}$ are the raw images. We aim to propagate the reliable estimates in low-textured areas from coarser scales to help with the estimation of finer scales without much loss in details.

We first use our basic MVS model with photometric consistency, ACMH, to obtain the initial depth maps for all images at the coarsest scale. To enhance the coherence among all depth maps, we further perform ACMH with geometric consistency. Then we upsample the depth maps to the next scale. The upsampling propagates the reliable depth estimates in low-textured areas to the current scale, which are obtained at the previous scale. To correct the errors induced from the previous scale, a detail restorer is first employed. These corrected depth maps are utilized as initialization to guide the subsequent ACMH with geometric consistency such that the reliable estimates within low-textured areas can be kept and optimized at the current scale. The same upsampling, detail restorer and ACMH with geometric consistency are repeated until we obtain the depth maps at the original image scale. We term our whole method ACMM.

\section{Structured Region Information}\label{Sec:ACMH}

\emph{Structured region information} means that pixels within a relatively large region can be approximately be modeled by the same 3D plane. Our basic MVS method with Adaptive Checkerboard sampling and Multi-Hypothesis joint view selection (ACMH) is inspired by this to sample better candidate hypotheses for propagation and select views with more credibility for multi-view matching costs aggregation. The details of ACMH are given as follows.

\subsection{Random Initialization} 

Following \cite{Galliani2015Massively}, we first randomly generate a hypothesis (including depth and normal) to build a 3D plane for each pixel in the reference image $I_{\text{ref}}$. For each hypothesis, a matching cost is computed from each of $N-1$ source images via a plane-induced homography~\cite{Hartley2004Multiple}. Then the top $K$ best matching costs are aggregated into the initial multi-view matching cost for the subsequent propagation. 

\subsection{Adaptive Checkerboard Sampling} 

We first adopt the idea in \cite{Galliani2015Massively} to partition the pixels of $I_{\text{ref}}$ into red-black grids of a checkerboard. This pattern allows us to simultaneously update the hypotheses of black pixels using red pixels and vice versa. In \cite{Galliani2015Massively}, their method samples from eight fixed positions. Differently, for each pixel in red or black group, we expand these eight points into four V-shaped areas and four long strip areas (Figure~\ref{fig:subfig:c}). Each V-shaped area contains $7$ samples while every long strip area contains $11$ samples. Then we sample eight good hypotheses from these areas according to their previous multi-view matching costs. This sampling scheme is favored by the structured region information. It means that a hypothesis with a smaller multi-view matching cost will represent a local plane better. This strategy helps a good plane of a local shared region to spread further as much as possible and supplies more compact estimates.

\subsection{Multi-Hypothesis Joint View Selection}

To obtain a robust multi-view matching cost for each pixel, we further leverage these eight structured hypotheses to infer the weight of every neighboring views. For pixel $p$, we calculate its corresponding matching costs with propagated hypotheses and embed them into a cost matrix     
\begin{equation}
\setlength{\abovedisplayskip}{2pt}
\setlength{\belowdisplayskip}{2pt}
{\textsf M}=
\begin{bmatrix}
m_{1,1} & m_{1,2} & \cdots & m_{1,N-1} \\
m_{2,1} & m_{2,2} & \cdots & m_{2,N-1} \\
\vdots & \vdots & \ddots & \vdots \\
m_{8,1} & m_{8,2} & \cdots & m_{8,N-1}
\end{bmatrix},
\end{equation}
where $m_{i,j}$ is the matching cost for the ${i}$-th hypothesis $h_{i}$ scored by the ${j}$-th view $I_{j}$. We adopt the bilateral weighted adaption of normalized cross correlation~\cite{Schonberger2016Pixelwise} to compute the matching cost, which describes the \emph{photometric consistency} between the reference and source patch. 

To infer aggregation views from the above cost matrix, we apply a voting decision in each column to determine whether a view is appropriate. A key observation behind this is that for a bad view, its corresponding eight matching costs are always high. In contrast, a good view always has some smaller matching costs. Furthermore, the matching costs for the
good view will decrease with the iteration of our algorithm. Therefore, a good matching cost boundary is defined as
\begin{equation}
\setlength{\abovedisplayskip}{2pt}
\setlength{\belowdisplayskip}{2pt}
{\tau}(t)={\tau}_{0}\cdot e^{-\frac{t^2}{\alpha}},
\end{equation}
where ${t}$ means the ${t}$-th iteration, ${\tau}_{0}$ is the initial matching cost threshold and $\alpha$ is a constant. Besides, we define a fixed bad matching cost threshold $\tau_{1}$ ($\tau_{1}>\tau(t)$). Based on our above observation, for a specific view $I_{j}$,  there should exist more than $n_{1}$ matching costs meeting the condition: $m_{i,j}<\tau(t)$. We define this set as $\textrm{S}_{\text{good}}(j)$ to calculate the weight of view $I_{j}$ later. Also, there should be less than $n_{2}$ matching costs meeting the condition: $m_{i,j}>\tau_{1}$. A view simultaneously satisfying the above conditions will be incorporated into the current view selection set $\textrm{S}_{t}$ in the $t$-th iteration.

The above inferred view selection set $\textrm{S}_{t}$ may contain some unstable views because of noise, viewing point and scale, \emph{etc}. This means each selected view will contribute different weights to the final aggregated matching cost. To evaluate the importance of each selected view, the confidence of a matching cost is computed as follows,
\begin{equation}
\setlength{\abovedisplayskip}{2pt}
\setlength{\belowdisplayskip}{2pt}
C({m}_{ij})=e^{-\frac{{m}_{ij}^2}{2{\beta}^2}}.
\end{equation}
where $\beta$ is a constant. This makes good views more discriminative. The weight of each selected view can be defined as
\begin{equation}\label{eq:1}
\setlength{\abovedisplayskip}{2pt}
\setlength{\belowdisplayskip}{2pt}
w({I}_{j}) = \frac{1}{|\textrm{S}_{\text{good}}(j)|}{\sum_{m_{i,j}\in\textrm{S}_{\text{good}}(j)}{C(m_{i,j})}}, {I}_{j}\in\textrm{S}_{t}.
\end{equation}
We suppose the most important view $v_{t-1}$ in iteration $t-1$ shall continue to have influence on the view selection of the current iteration $t$. Thus, we modify Formula~\ref{eq:1} as
\begin{equation}
\setlength{\abovedisplayskip}{2pt}
\setlength{\belowdisplayskip}{2pt}
w'({I}_{j})=\left\{ \begin{array}{ll}
(\mathbb{I}({I}_{j}={v}_{t-1})+1)\cdot{w}({I}_{j}), & \textrm{if}\;{I}_{j}\in\textrm{S}_{t}; \\
0.2\cdot{\mathbb{I}({I}_{j}={v}_{t-1})}, & {\textrm{else.}}
\end{array} \right.
\end{equation}
where $\mathbb{I}(\cdot)$ is an indicator function such that $\mathbb{I}(\text{true})=1$ and $\mathbb{I}(\text{false})=0$. This modification can make our view selection method more robust. With the inferred weights $w'$, the multi-view aggregated photometric consistency cost of pixel $p$ for hypothesis $h_{i}$ is defined as 
\begin{equation}\label{eq:2}
\setlength{\abovedisplayskip}{2pt}
\setlength{\belowdisplayskip}{2pt}
m_{\text{photo}}(p,h_{i})=\frac{\sum_{j=1}^{N-1}w'({I}_{j})\cdot{m_{i,j}}}{\sum_{j=1}^{N-1}w'({I}_{j})}.
\end{equation}
The current best estimate for pixel $p$ is updated by the hypothesis with the minimum multi-view aggregated cost.

\subsection{Refinement} 

After each red-black iteration, a refinement step is applied to enrich the diversity of solution space. There exist three conditions for the current depth and normal of pixel $p$, \ie, either of them, neither of them, or both of them are close to the optimal solution~\cite{Schonberger2016Pixelwise}. Thus, we generate two new hypotheses, one of which is randomly generated and the other is obtained by perturbing the current estimate. We combine these new depths and normals with the current depth and normal, yielding another six new hypotheses to be tested. The hypothesis with the least aggregated cost is chosen as the final estimate for pixel $p$. The above propagation, view selection and refinement are repeated multiple times to get the final depth map for $I_{\text{ref}}$. At the end, a median filter of size $5\times5$ is applied to our final depth maps.

\section{Multi-Scale Geometric Consistency}\label{Sec:PGC}

Combined with the multi-scale scheme, ACMH at the coarsest scale obtains more reliable depth estimates in low-textured areas. However, photometric consistency experiences difficulties when applied to optimize these depth estimates at finer scales. In this section, we detail how to leverage geometric consistency guidance to deal with the optimization of these estimates. Also, a detail restorer is present to correct the errors induced from coarser scales.

\subsection{Geometric Consistency Guidance}  

After obtaining the reliable depth estimates for low-texture areas at the coarsest scale and propagating them to finer scales via upsampling, we need to optimize these estimates at finer scales. Our key idea is that the upsampled depth maps of source images can geometrically constrain these estimates from being disturbed by photometric consistency, which means \emph{geometric consistency}. Inspired by \cite{Zhang2008Recovering,Schonberger2016Pixelwise}, we use the forward-backward reprojection error to indicate this consistency. 

Given the depth of pixel $p$ in image $I_{i}$ is known as $D_{i}(p)$, with the camera parameter $P_{i}=[M_{i}|{\rm p}_{i,4}]$~\cite{Hartley2004Multiple}, its corresponding back-projected 3D point $X_{i}(p)$ is computed as 
\begin{equation}
\setlength{\abovedisplayskip}{2pt}
\setlength{\belowdisplayskip}{2pt}
X_{i}(p)=M_{i}^{-1}\cdot(D_{i}(p)\cdot{p}-{\rm p}_{i,4}).
\end{equation}
Then the reprojection error between the reference image $I_{\text{ref}}$ and the source image $I_{j}$ for $i$-th hypothesis is calculated as 
\begin{equation}\label{eq:4}
\setlength{\abovedisplayskip}{2pt}
\setlength{\belowdisplayskip}{2pt}
\Delta{e}_{i,j}={\text{min}}(\|P_{\text{ref}}\cdot{X_{j}(P_{j}\cdot{X_{\text{ref}}(p)})}-p\|,\delta),
\end{equation}
where $\delta$ is a truncation threshold to robustify the reprojection error against occlusions. We integrate the above equation into Formula~\ref{eq:2} and get the following multi-view aggregated geometric consistency cost as
\begin{equation}\label{eq:5}
\setlength{\abovedisplayskip}{2pt}
\setlength{\belowdisplayskip}{2pt}
m_{\text{geo}}(p,h_{i})=\frac{\sum_{j=1}^{N-1}w'({I}_{j})\cdot({m}_{i,j}+\lambda\cdot\Delta{e}_{i,j})}{\sum_{j=1}^{N-1}w'({I}_{j})},
\end{equation}
where $\lambda$ is a factor that balances the weight of the two terms. 

Specifically, at the $l$-th scale ($l>0$), we employ the joint bilateral upsampler~\cite{Kopf2007JBU} to propagate the estimates at the previous scale to the current scale. The upsampled estimates are utilized as the initial seeds of the current scale to perform the subsequent propagation, view selection and refinement as in ACMH. Differently, here we adopt Formula~\ref{eq:5} instead of Formula~\ref{eq:2} to update the hypothesis of pixel $p$. In fact, this modification limits the solution space of current hypothesis update, especially for the hypothesis update in low-textured areas. This guarantees that the reliable estimates in low-textured areas obtained at the coarsest scale can be propagated to the finest scale. It is worth noting that the geometric consistency also optimizes the depth estimation of other areas except low-textured areas.

Additionally, we notice that the initial depth maps obtained by ACMH are noisy due to ambiguities and occlusions. However, photometric consistency is hard to reflect these errors since large depth variations only induce small cost changes~\cite{Schonberger2016Pixelwise}. Thus, we also perform geometric consistency at the coarsest scale to optimize these initial depth maps. Intuitively, if the neighboring depth maps are estimated more accurately, the depth map of the reference image will be further boosted. Thus, we conduct geometric consistency guidance twice to refine depth maps at each scale in our experiments. 

\begin{figure*}[t]
	\setlength{\abovecaptionskip}{0pt}
	\setlength{\belowcaptionskip}{0pt}
	\centering
	\subfigure[]{
		\label{fig:subfig:DPR_a} 
		\includegraphics[width=0.157\linewidth]{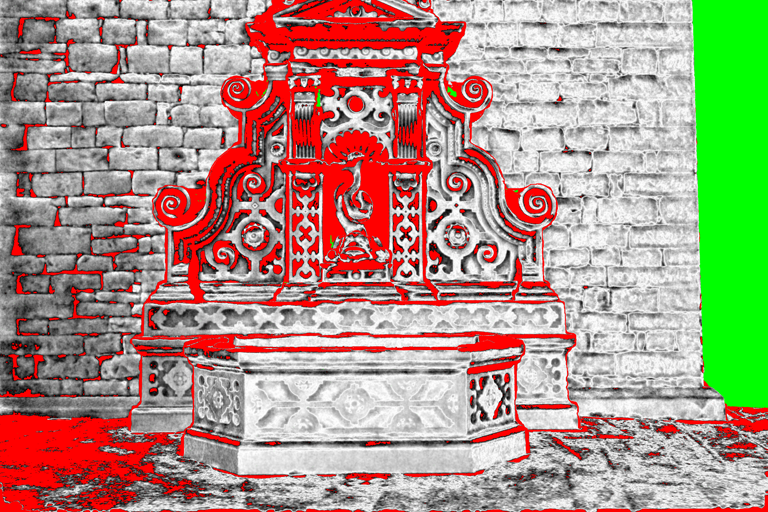}} 
	\subfigure[]{
		\label{fig:subfig:DPR_b}
		\includegraphics[width=0.157\linewidth]{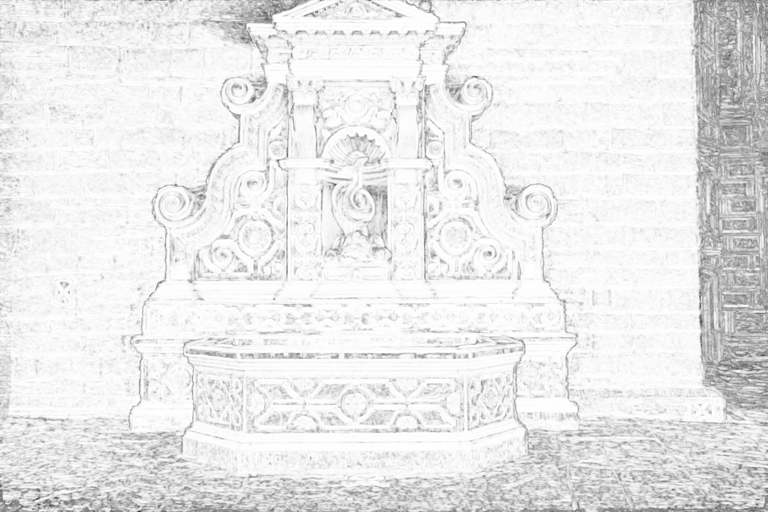}}
	\subfigure[]{
		\label{fig:subfig:DPR_c}
		\includegraphics[width=0.157\linewidth]{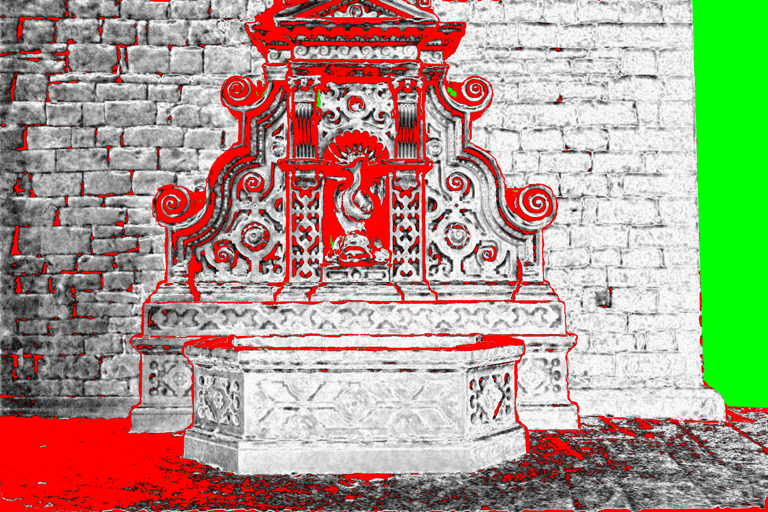}}
	\subfigure[]{
		\label{fig:subfig:DPR_d}
		\includegraphics[width=0.157\linewidth]{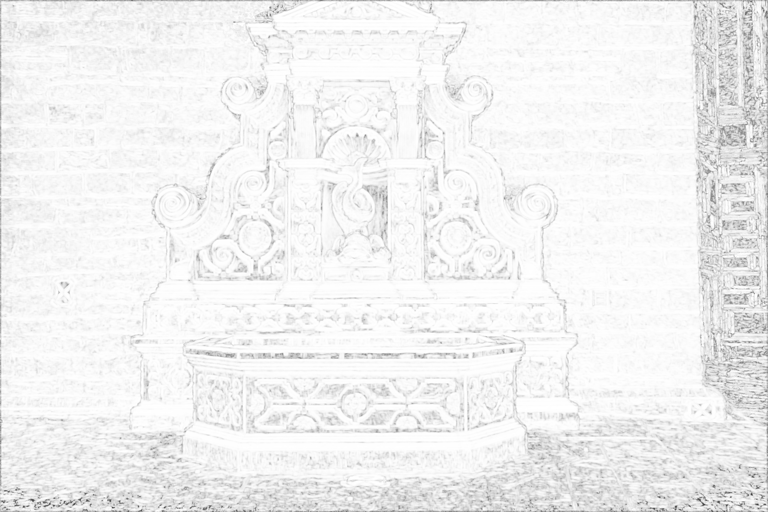}}
	\subfigure[]{
		\label{fig:subfig:DPR_e}
		\includegraphics[width=0.157\linewidth]{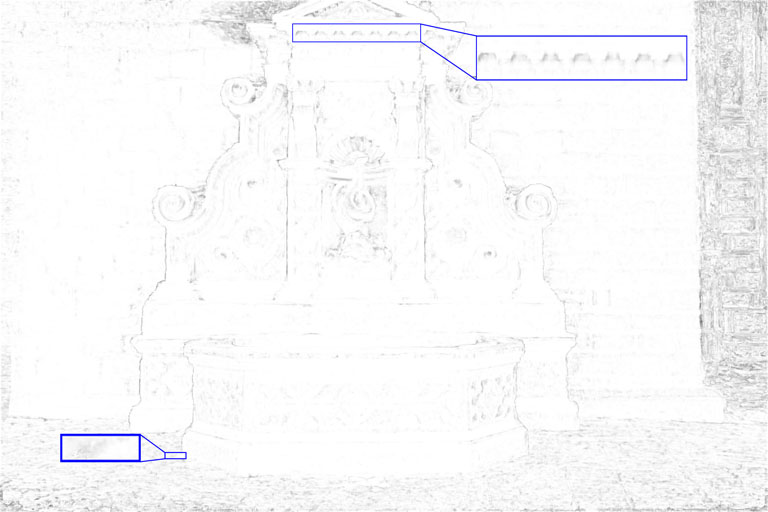}}
	\subfigure[]{
		\label{fig:subfig:DPR_f}
		\includegraphics[width=0.157\linewidth]{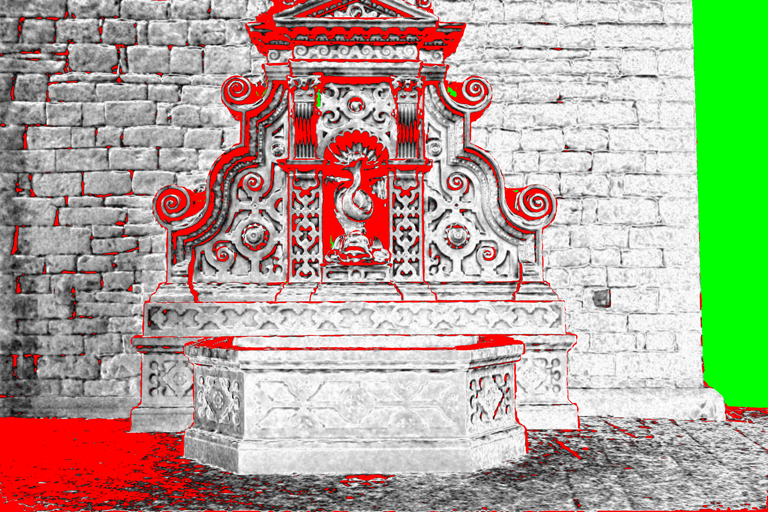}}
	\caption{Absolute error maps and photometric consistency cost maps on Fountain-P11 dataset for different methods. (a) shows the absolute error map of our method without detail restorer. Its depth map is obtained by upscaling the estimation of the penultimate scale. Details are not preserved. (b) shows the photometric consistency cost map of (a). (c) shows the absolute error map of our basic MVS model. Its details are better preserved than (a). (d) shows the photometric consistency cost map of (c). (e) shows the difference map of (b) and (d). The cost difference of the erroneous estimates in details is more discriminative than the cost difference of the reliable estimation in low-textured areas. (f) shows the absolute error map of ACMM. For absolute error maps, green pixels encode missing ground truth data, red pixels encode an absolute error larger than $2cm$, and pixels with absolute errors between 0 and $2cm$ are encoded in gray [255, 0].}
	\label{fig:DPR}
\end{figure*}

\subsection{Detail Restorer} 

The multi-scale geometric consistency guidance on the one hand helps with the estimation of low-textured areas but on the other hand often leads to blurred details. At the coarser scales, the lost image details directly cause the loss of their depth information. Additionally, the fixed patch window size makes ACMH hard to achieve a trade-off between thin structures and low-textured areas because the local planar assumption does not hold for details~\cite{Kanade1994Stereo,Zhang2009Cross,Yoon2005Locally}. Furthermore, although upsampling can spread the reliable estimates in low-textured areas to larger regions, it also brings some extra errors in details. However, we observe that these details can be better estimated at the original image scale with only photometric consistency (Figure~\ref{fig:subfig:DPR_c}). Thus, we consider how to leverage photometric consistency to probe erroneous estimates in details and correct them.

As shown in Figure~\ref{fig:subfig:DPR_a}, the blurred details often happen in thin structures or boundaries. We hope to detect these regions and only enforce photometric consistency in these specific regions to rectify the erroneous estimates. We observe that the difference map of photometric consistency cost between adjacent scales can magnify the errors in details while suppressing the reflection of reliable estimates in low-textured areas (Figure~\ref{fig:subfig:DPR_e}). Thus, we can leverage this difference map to probe the errors in details and correct them in a unified way. Specifically, after we upsample the estimates (\ie, depth and normal) of the previous scale, we use them to recompute the photometric consistency cost $C^{l}_{\text{init}}$ at the current scale. Then, we execute the basic MVS model to get new photometric consistency cost $C^{l}_{\text{photo}}$. The estimate for pixel $p$ will be considered as an error if the difference of photometric consistency cost fulfills 
\begin{equation}
\setlength{\abovedisplayskip}{2pt}
\setlength{\belowdisplayskip}{2pt}
C^{l}_{\text{init}}(p)-C^{l}_{\text{photo}}(p)>\xi,
\end{equation}  
where $\xi$ is a small constant value that increases the robustness to distinguish the erroneous estimates. Meanwhile, the erroneous estimates will be replaced by the hypotheses reflecting the above difference. By combining the detail restorer, ACMM can make a better trade-off between low-textured areas and details as shown in Figure~\ref{fig:subfig:DPR_f}.


\section{Fusion} 

After getting all depth maps, we adopt a fusion step similar to \cite{Galliani2015Massively,Schonberger2016Pixelwise} to merge them into a complete point cloud. Specifically, we cast each image as reference image in turn, convert its depth map to 3D points in world coordinate and project them to its neighboring views to get corresponding matches. We define a consistent match satisfying the relative depth difference $\epsilon\leq0.01$, the angle between normals $\theta\leq30^\circ$ and the reprojection error $\psi\leq2$ as in \cite{Schonberger2016Pixelwise}. If there exist $n\geq2$ neighboring views whose corresponding matches satisfy the above constraints, the depth estimate will be accept. At last, the 3D points and normal estimates corresponding to these consistent depth estimates are averaged into a unified 3D point.

\section{Experiments}\label{Sec:Exp}

\begin{table*}[t]
	\caption{Percentage of pixels with absolute errors below $2cm$ and $10cm$ on Strecha dataset. The related values are from \cite{Hu2012Least,Zheng2014PatchMatch,Schonberger2016Pixelwise}. \cite{Schonberger2016Pixelwise}$\backslash$G means COLMAP without geometric consistency.}
	\centering
	\footnotesize
	\begin{tabular}{c|c|c|c|c|c|c|c|c|c|c|c|c|c}
		\hline
		& error & \cite{Zheng2014PatchMatch} & \cite{Zach2008Fast} & \cite{Furukawa2010Accurate} & \cite{Zaharescu2011Topology} & \cite{Tylecek2010Refinement} & \cite{Jancosek2011Multi} & \cite{Galliani2015Massively} & \cite{Schonberger2016Pixelwise} & \cite{Schonberger2016Pixelwise}$\backslash$G & DWTA & {ACMH}  & {ACMM} \\
		\hline
		\multirow{2}{*}{Fountain} & $2 cm$ & 0.769 & 0.754 & 0.731 & 0.712 & 0.732 & 0.824 & 0.693 & 0.827 & 0.804 & 0.778 & 0.793 & \bf{0.853} \\
		& $10 cm$ & 0.929 & 0.930 & 0.838 & 0.832 & 0.822 & 0.973 & 0.838 & \bf{0.975} & 0.949 & 0.921 & 0.952 & 0.974 \\
		\hline
		\multirow{2}{*}{HerzJesu} & $2 cm$ & 0.650 & 0.649 & 0.646 & 0.220 & 0.658 & \bf{0.739} & 0.283 & 0.691 & 0.679 & 0.614 & 0.656 & 0.731 \\
		& $10 cm$ & 0.844 & 0.848 & 0.836 & 0.501 & 0.852 & 0.923 & 0.455 & 0.931 & 0.907 & 0.804 & 0.873 & \bf{0.932} \\
		\hline
	\end{tabular}
	\label{tab:S-depthmap}
\end{table*}

\begin{table*}[t]
	\caption{Percentage of pixels with absolute errors below $2cm$ and $10cm$ on the high-resolution multi-view training datasets of ETH3D benchmark. The best results are marked in bold while the second-best results are marked in red.}
	\centering
	\footnotesize
	\begin{tabular}{c|c|c|c|c|c|c|c|c|c|c|c|c|c|c}
		\hline
		\multirow{2}{*}{error} & \multirow{2}{*}{method} & \multicolumn{7}{c|}{indoor} & \multicolumn{6}{c}{outdoor} \\ 
		\cline{3-15}
		& & delive. & kicker & office & pipes & relief & relief. & terrai. & courty. & electro & facade & meadow & playgr. & terrace \\
		\hline
		\multirow{4}{*}{$2 cm$} & \cite{Schonberger2016Pixelwise} & 0.697 & \textcolor{red}{0.435} & 0.263 & 0.411 & 0.863 & 0.858 & 0.576 & \textcolor{red}{0.826} & 0.710 & \textcolor{red}{0.742} & 0.546 & 0.709 & 0.808 \\
		& DWTA & 0.705 & 0.369 & 0.293 & 0.419 & 0.887 & 0.883 & 0.675 & 0.772 & 0.730 & 0.684 & 0.464 & 0.731 & 0.801 \\
		\cline{2-15}
		& ACMH & \textcolor{red}{0.733} & 0.427 & \textcolor{red}{0.323} & \textcolor{red}{0.536} & \textcolor{red}{0.891} & \textcolor{red}{0.903} & \textcolor{red}{0.714} & 0.799 & \textcolor{red}{0.748} & 0.685 & \textcolor{red}{0.571} & \textcolor{red}{0.753} & \textcolor{red}{0.820} \\
		& ACMM & \bf{0.777} & \bf{0.667} & \bf{0.512} & \bf{0.765} & \bf{0.960} & \bf{0.957} & \bf{0.854} & \bf{0.844} & \bf{0.868} & \bf{0.745} & \bf{0.771} & \bf{0.843} & \bf{0.897} \\
		\hline
		\multirow{4}{*}{$10 cm$} & \cite{Schonberger2016Pixelwise} & 0.806 & 0.514 & 0.342 & 0.478 & 0.896 & 0.893 & 0.635 & 0.934 & 0.774 & \textcolor{red}{0.909} & 0.701 & 0.810 & 0.891 \\
		& DWTA & 0.815 & 0.451 & 0.382 & 0.496 & 0.918 & 0.918 & 0.738 & 0.910 & 0.810 & 0.899 & 0.647 & 0.844 & 0.894 \\
		\cline{2-15}
		& ACMH & \textcolor{red}{0.842} & \textcolor{red}{0.519} & \textcolor{red}{0.418} & \textcolor{red}{0.617} & \textcolor{red}{0.923} & \textcolor{red}{0.941} & \textcolor{red}{0.778} & \textcolor{red}{0.937} & \textcolor{red}{0.834} & 0.908 & \textcolor{red}{0.786} & \textcolor{red}{0.869} & \textcolor{red}{0.915} \\
		& ACMM & \bf{0.930} & \bf{0.800} & \bf{0.648} & \bf{0.839} & \bf{0.982} & \bf{0.984} & \bf{0.904} & \bf{0.973} & \bf{0.947} & \bf{0.934} & \bf{0.917} & \bf{0.951} & \bf{0.980} \\
		\hline
	\end{tabular}
	\label{tab:E-depthmap}
\end{table*}

\begin{figure*}[!t]
	\setlength{\abovecaptionskip}{0pt}
	\setlength{\belowcaptionskip}{0pt}
	\centering
	\subfigure[Reference image]{
		\begin{minipage}[t]{0.157\linewidth} 
			\centering
			\includegraphics[width=0.99\linewidth]{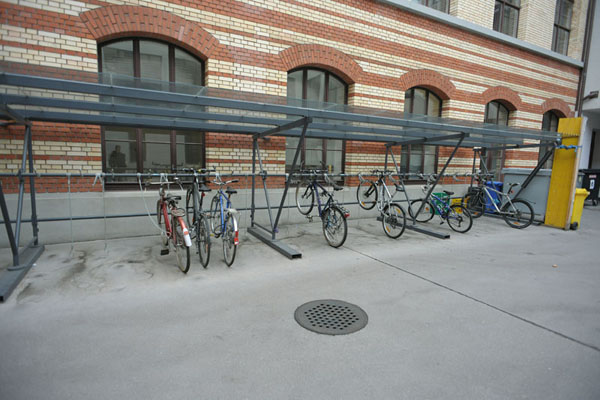}
			\includegraphics[width=0.99\linewidth]{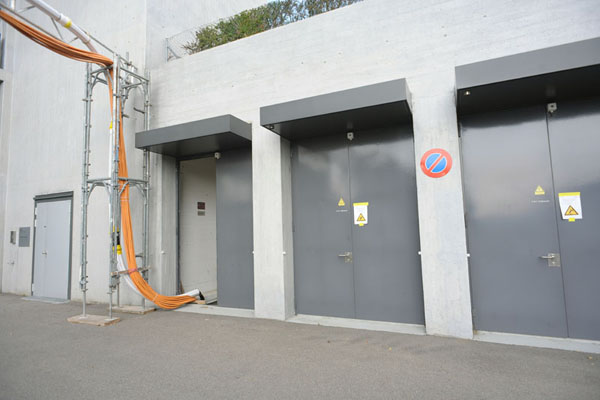}
			\includegraphics[width=0.99\linewidth]{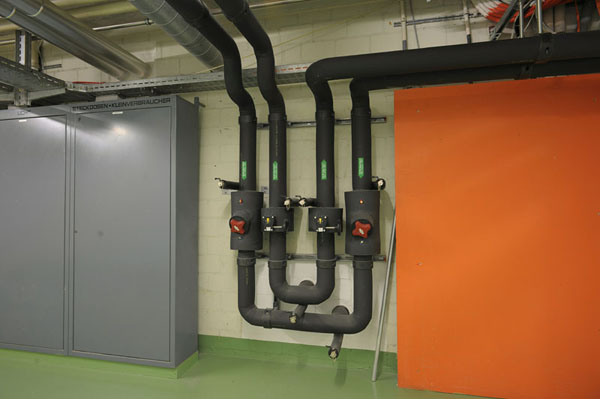}
	\end{minipage}}
	\subfigure[Ground truth]{
		\begin{minipage}[t]{0.157\linewidth}
			\centering
			\includegraphics[width=0.99\linewidth]{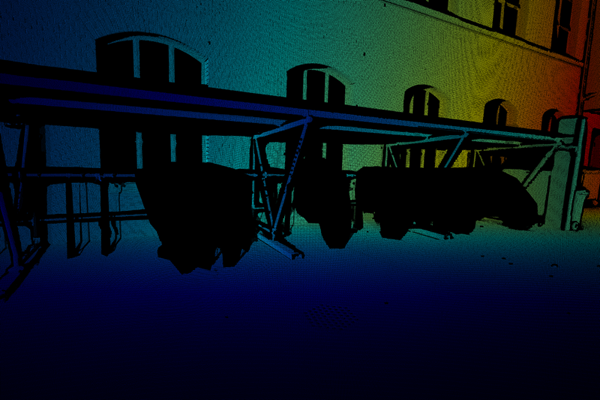}
			\includegraphics[width=0.99\linewidth]{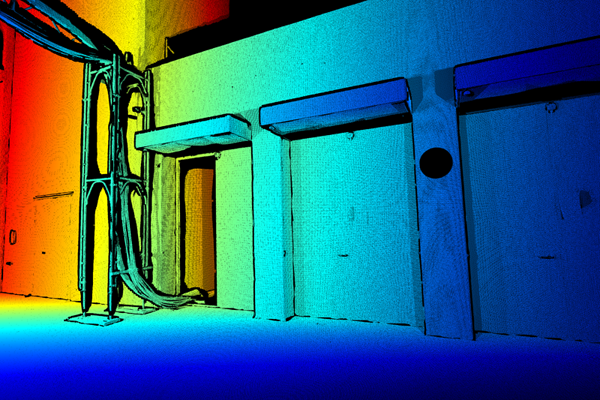}
			\includegraphics[width=0.99\linewidth]{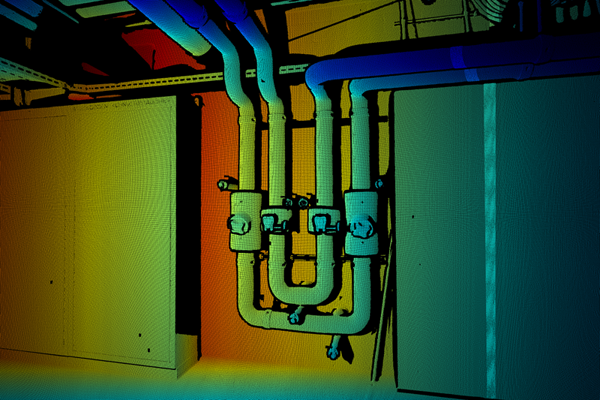}
	\end{minipage}}
	\subfigure[COLMAP]{
		\begin{minipage}[t]{0.157\linewidth}
			\centering
			\includegraphics[width=0.99\linewidth]{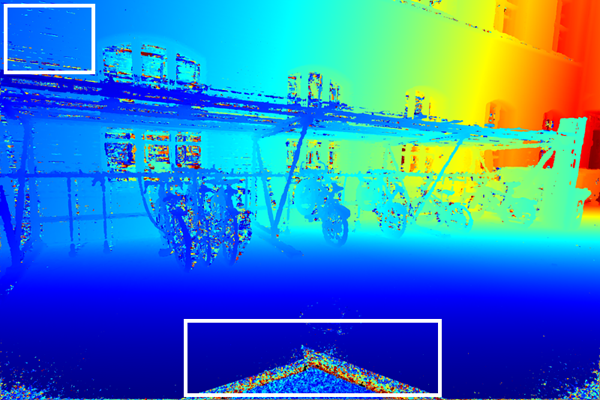}
			\includegraphics[width=0.99\linewidth]{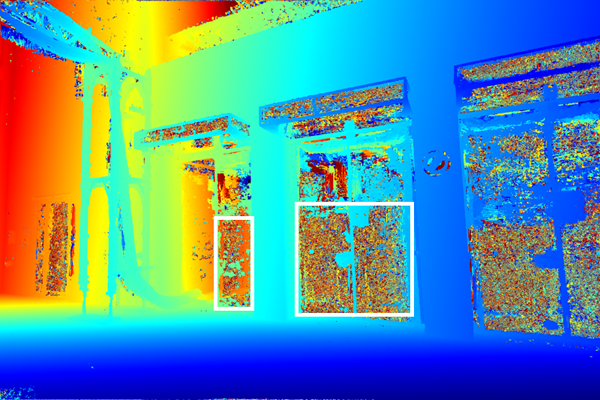}
			\includegraphics[width=0.99\linewidth]{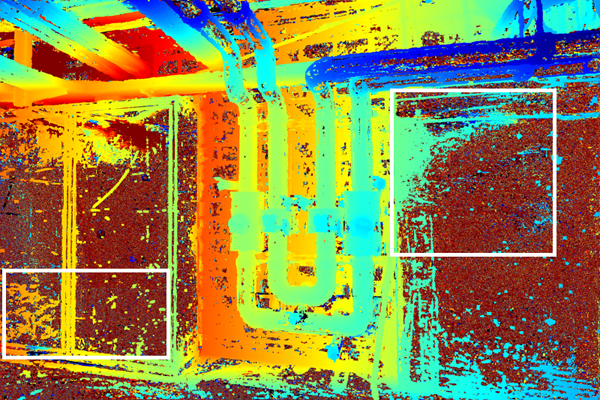}
	\end{minipage}}
	\subfigure[DWTA]{
		\begin{minipage}[t]{0.157\linewidth}
			\centering
			\includegraphics[width=0.99\linewidth]{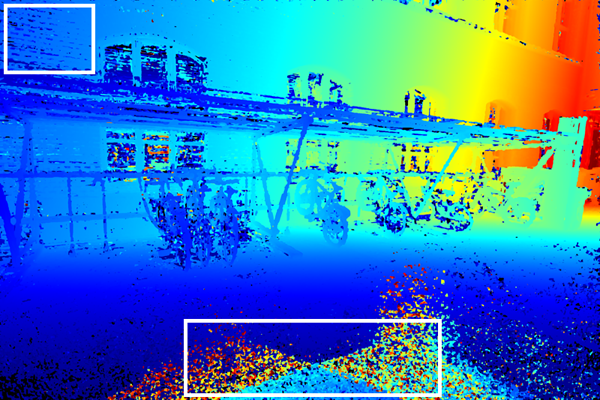}
			\includegraphics[width=0.99\linewidth]{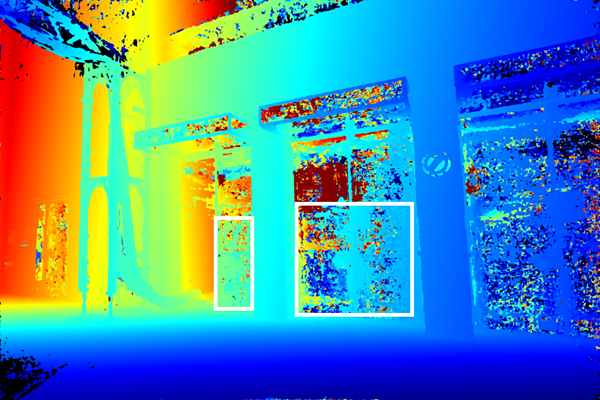}
			\includegraphics[width=0.99\linewidth]{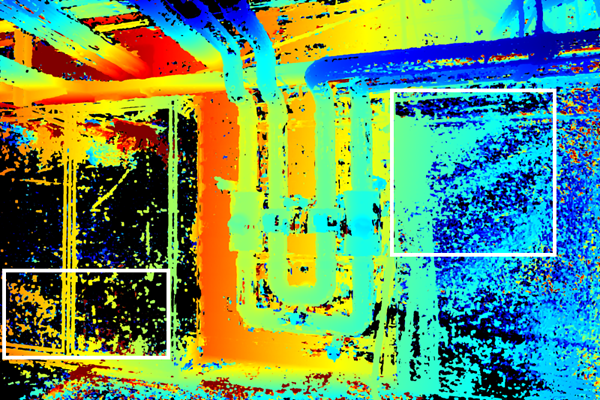}
	\end{minipage}}
	\subfigure[ACMH]{
		\begin{minipage}[t]{0.157\linewidth}
			\centering
			\includegraphics[width=0.99\linewidth]{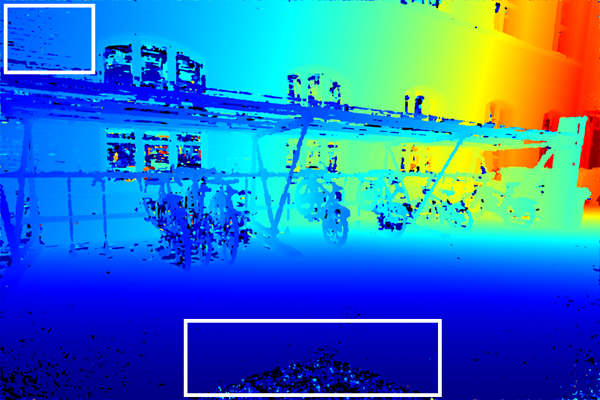}
			\includegraphics[width=0.99\linewidth]{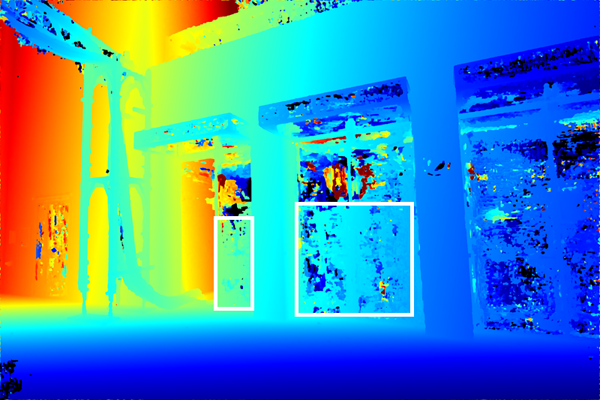}
			\includegraphics[width=0.99\linewidth]{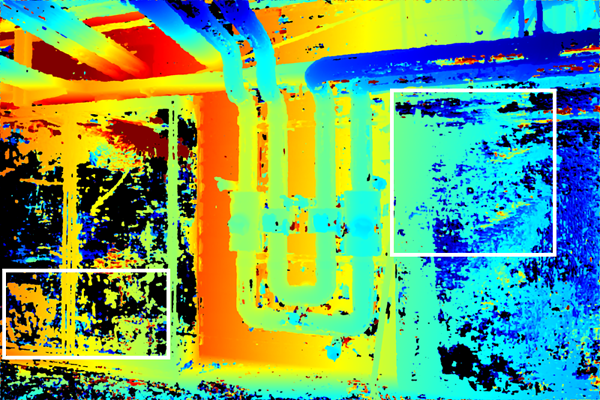}
	\end{minipage}}
	\subfigure[ACMM]{
		\begin{minipage}[t]{0.157\linewidth}
			\centering
			\includegraphics[width=0.99\linewidth]{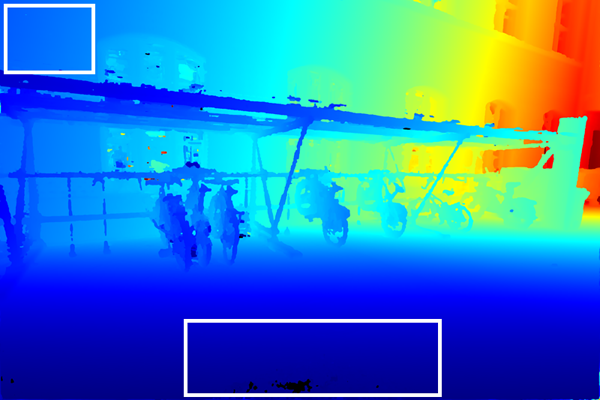}
			\includegraphics[width=0.99\linewidth]{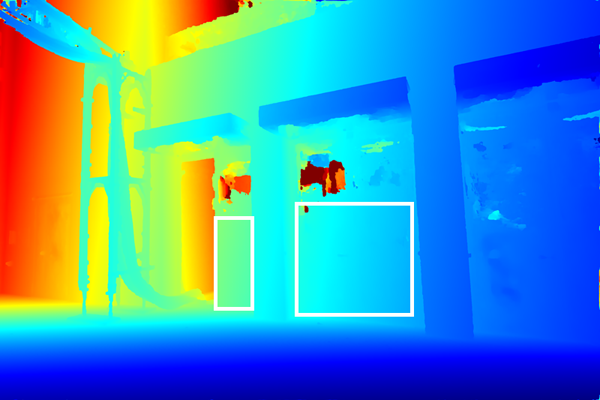}
			\includegraphics[width=0.99\linewidth]{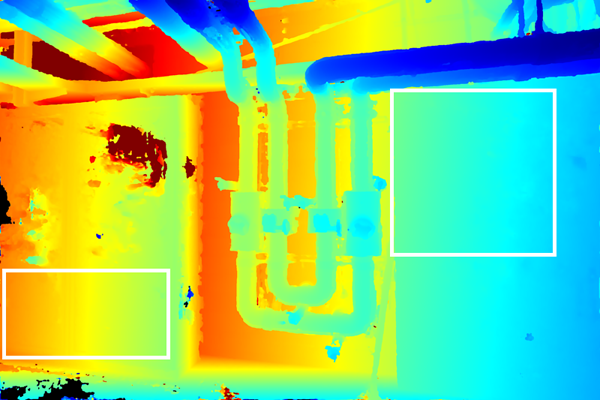}
	\end{minipage}}
	\caption{Qualitative depth map comparisons between different algorithms on some high-resolution multi-view training datasets (courty., electro, pipes) of ETH3D benchmark. Black pixels in (b) have no ground truth data. Some challenging areas are shown in white boxes. }
	\label{fig:depthmap}
\end{figure*}

\begin{figure*}[t]
	\setlength{\abovecaptionskip}{0pt}
	\setlength{\belowcaptionskip}{0pt}
	\centering
	\subfigure[Images]{
		\begin{minipage}[t]{0.148\linewidth} 
			\centering
			\includegraphics[width=0.99\linewidth]{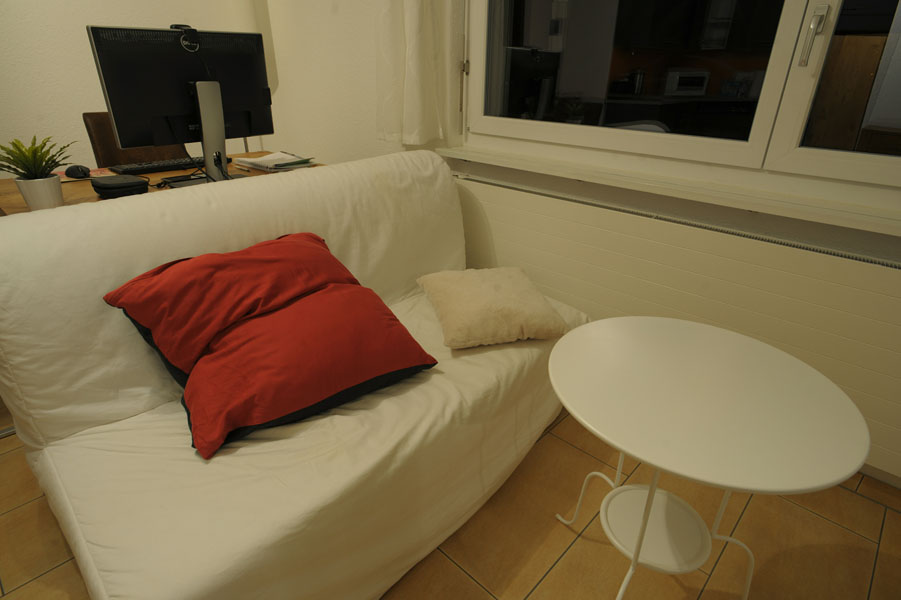}
			\includegraphics[width=0.99\linewidth]{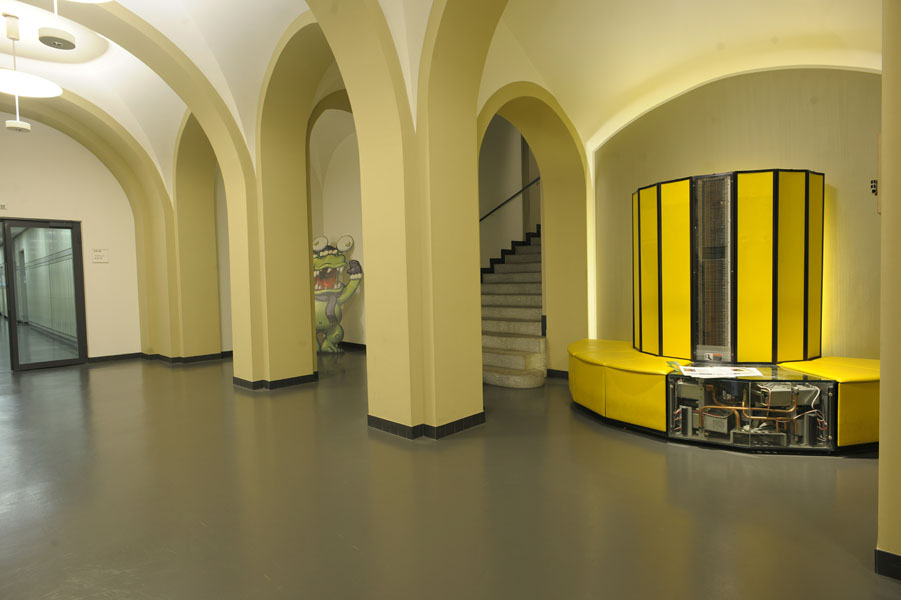}
	\end{minipage}}
	\subfigure[PMVS]{
		\begin{minipage}[t]{0.131\linewidth} 
			\centering
			\includegraphics[width=0.99\linewidth]{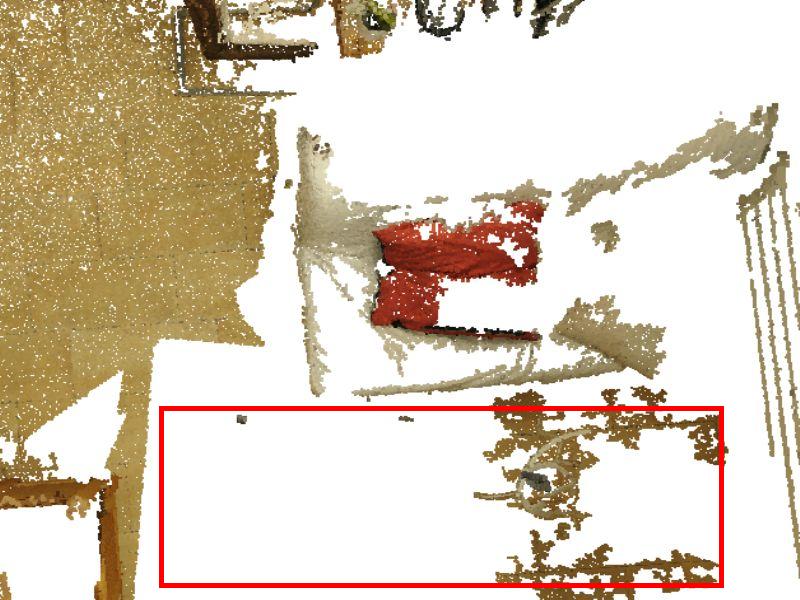}
			\includegraphics[width=0.99\linewidth]{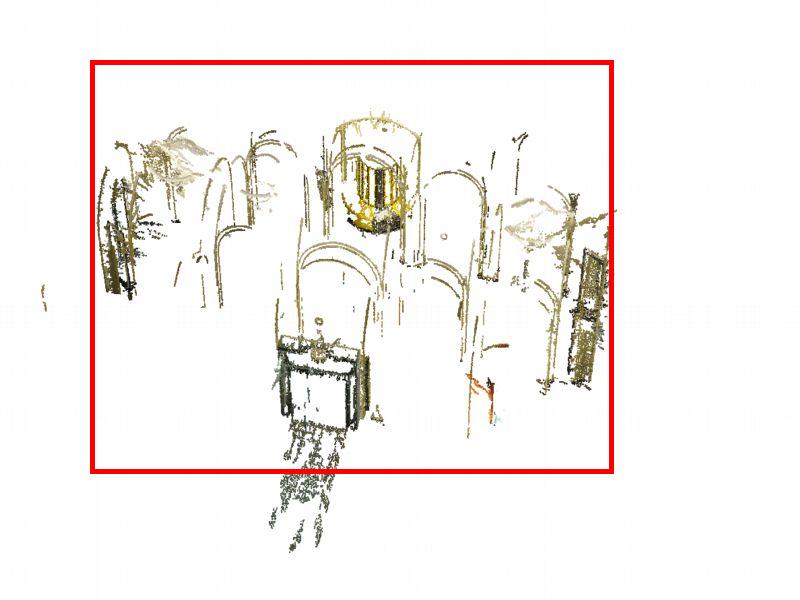}
	\end{minipage}}
	\subfigure[Gipuma]{
		\begin{minipage}[t]{0.131\linewidth}
			\centering
			\includegraphics[width=0.99\linewidth]{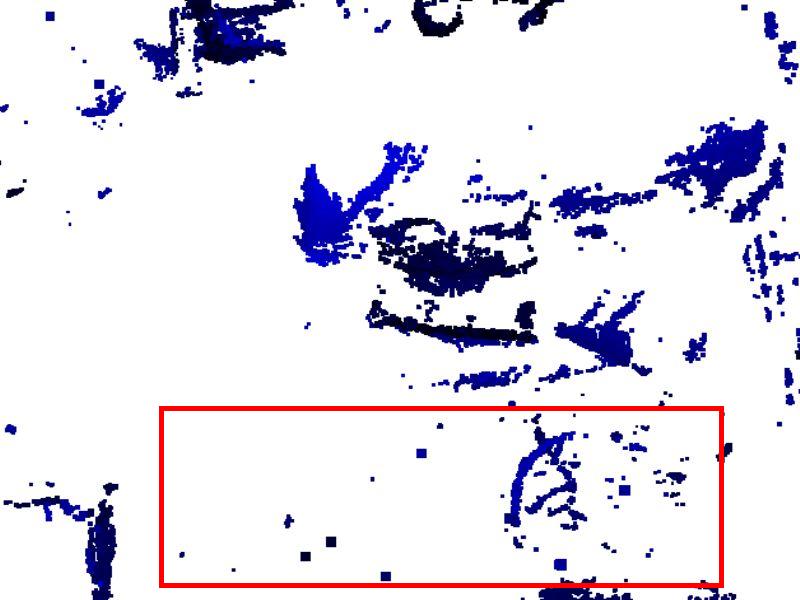}
			\includegraphics[width=0.99\linewidth]{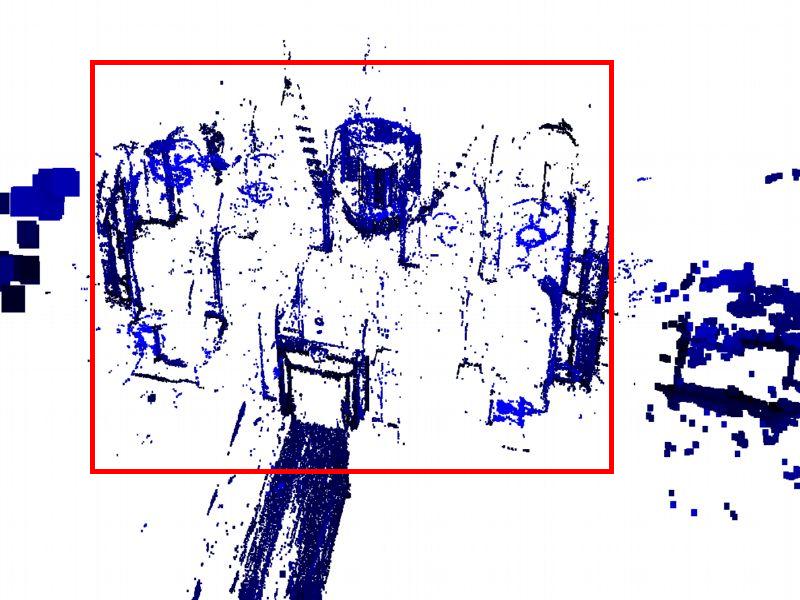}
	\end{minipage}}
	\subfigure[COLMAP]{
		\begin{minipage}[t]{0.131\linewidth}
			\centering
			\includegraphics[width=0.99\linewidth]{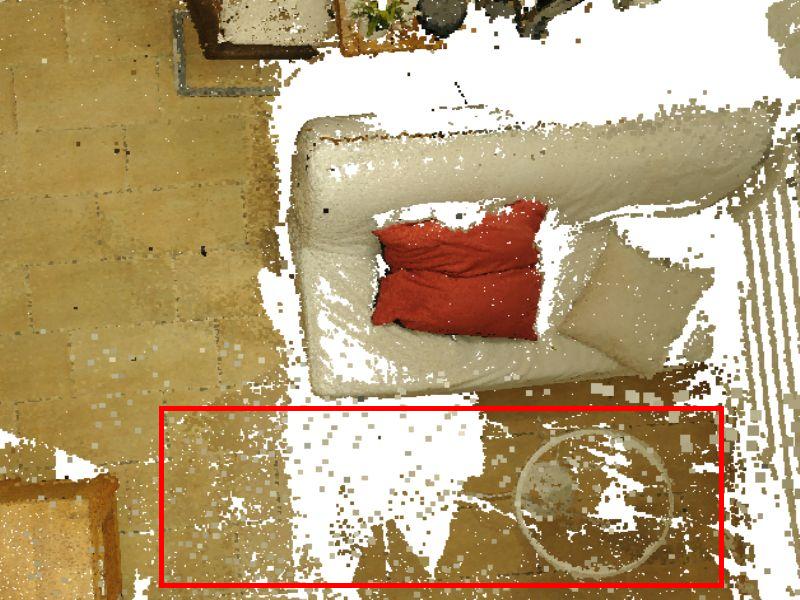}
			\includegraphics[width=0.99\linewidth]{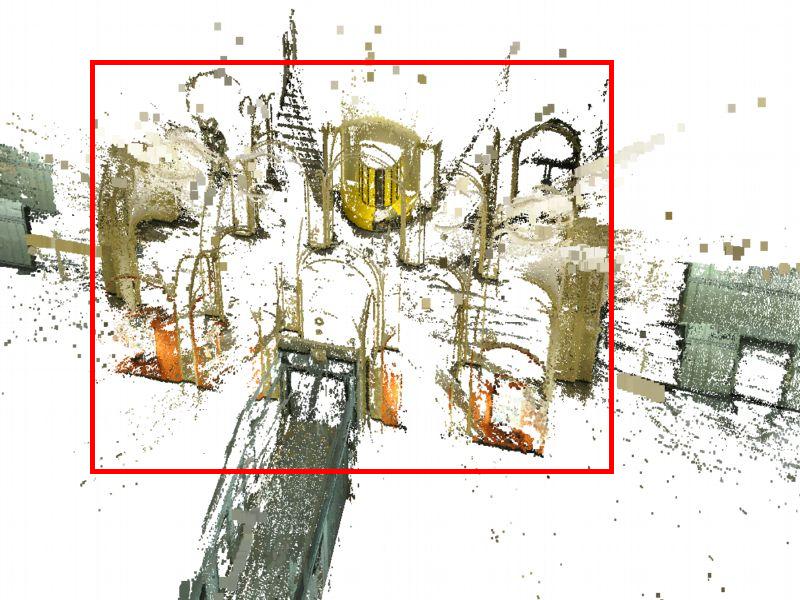}
	\end{minipage}}
	\subfigure[LTVRE]{
		\begin{minipage}[t]{0.131\linewidth}
			\centering
			\includegraphics[width=0.99\linewidth]{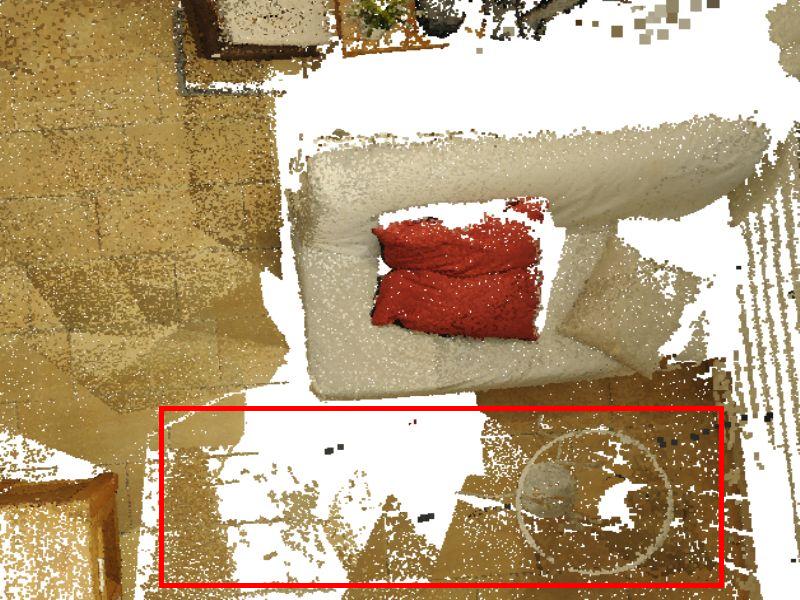}
			\includegraphics[width=0.99\linewidth]{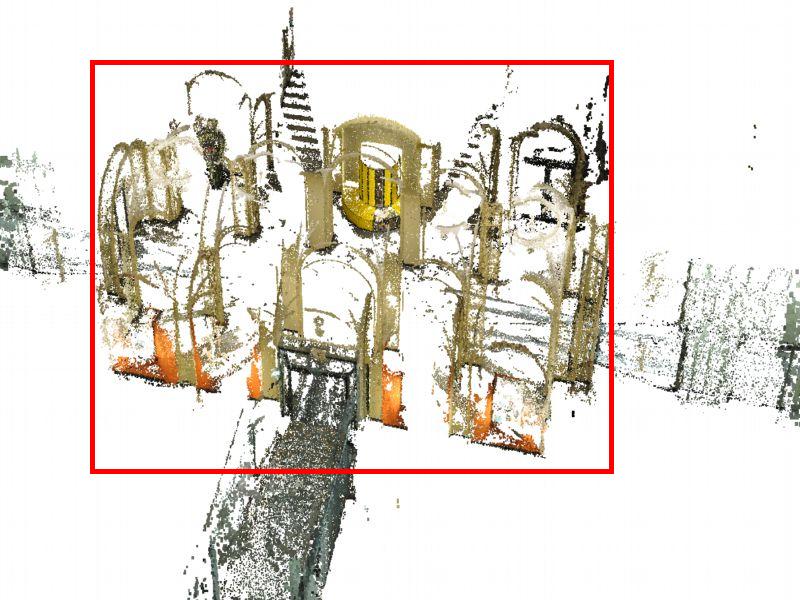}
	\end{minipage}}
	\subfigure[ACMH]{
		\begin{minipage}[t]{0.131\linewidth}
			\centering
			\includegraphics[width=0.99\linewidth]{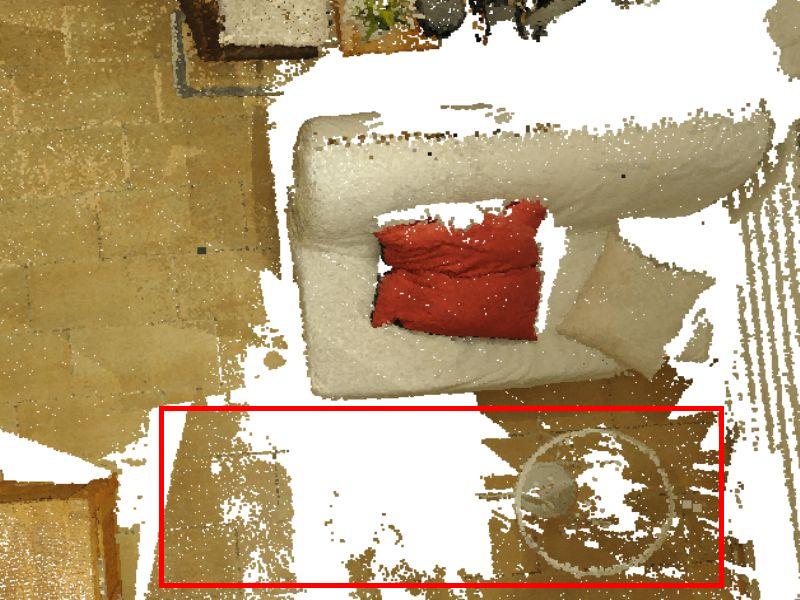}
			\includegraphics[width=0.99\linewidth]{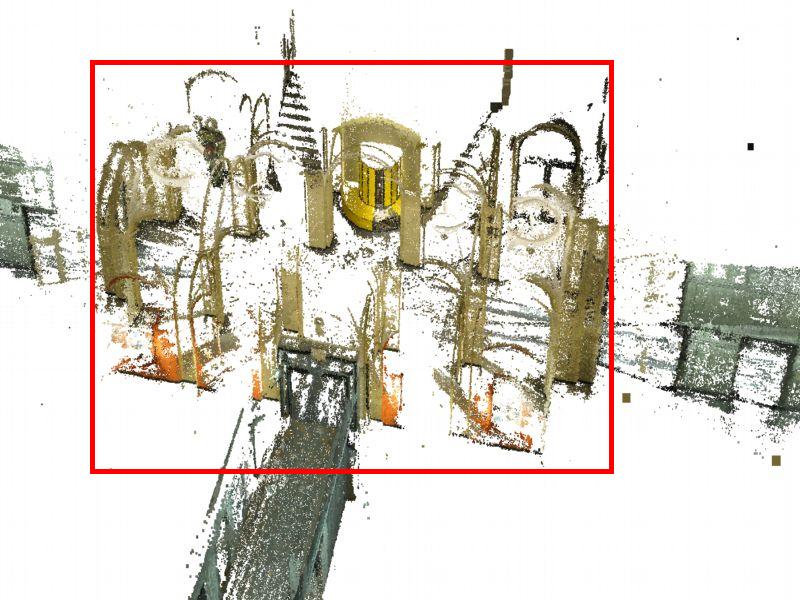}
	\end{minipage}}
	\subfigure[ACMM]{
		\begin{minipage}[t]{0.131\linewidth}
			\centering
			\includegraphics[width=0.99\linewidth]{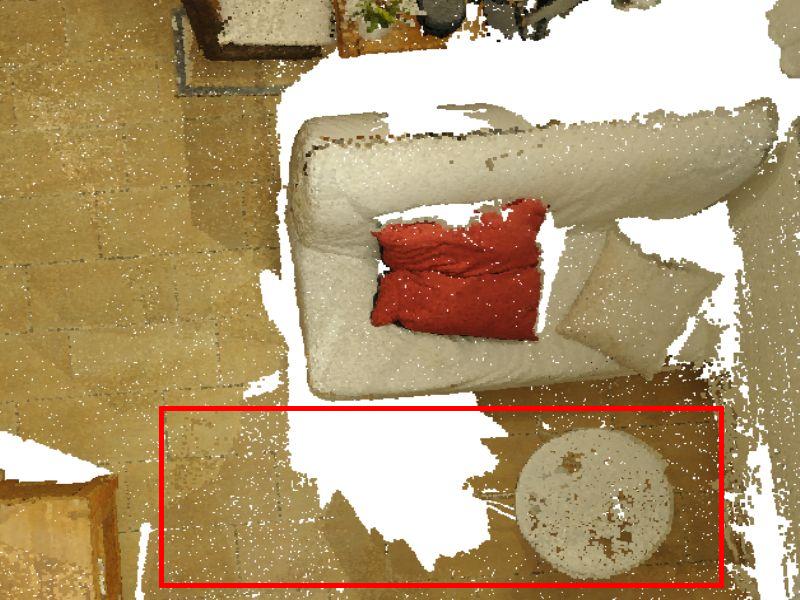}
			\includegraphics[width=0.99\linewidth]{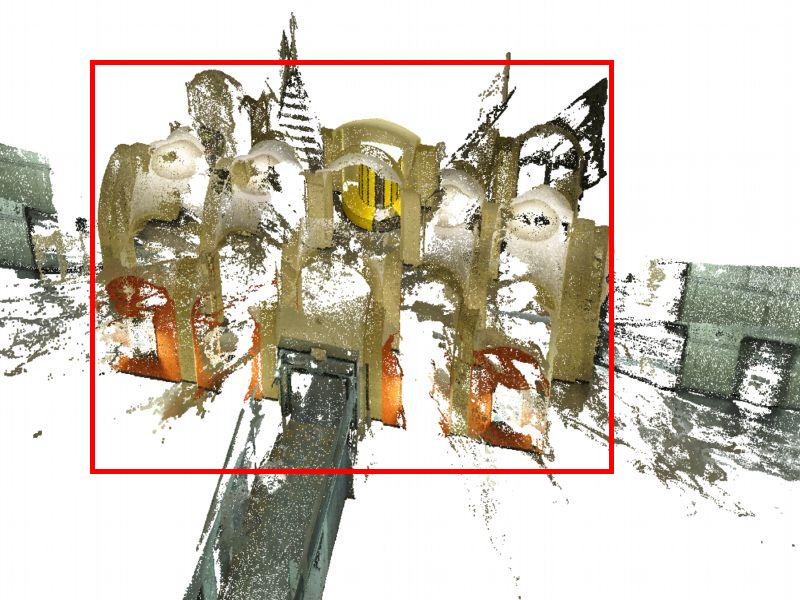}
	\end{minipage}}
	\caption{Qualitative point cloud comparisons between different algorithms on some high-resolution multi-view test datasets (living., old co.) of ETH3D benchmark. These dense 3D models are reported by the ETH3D benchmark evaluation server~\cite{Schops2017ETH3D}.}
	\label{fig:pointcloud_test}
\end{figure*}

We evaluate our method on two MVS datasets, Strecha dataset~\cite{Strecha2008On} and ETH3D benchmark~\cite{Schops2017Multi}, from two perspectives, depth map assessment and point cloud evaluation. 

\subsection{Datasets and Settings}

Strecha dataset~\cite{Strecha2008On} comprises two scenes with ground truth depth maps, Fountain and HerzJesu. They have $11$ and $8$ images respectively with $3072\times2048$ resolution. Although Strecha dataset provides relatively easy (\ie, well-textured) scenes and its online service is not available anymore, there are many state-of-the-art methods evaluating their depth maps on it. Thus, we will first utilize Strecha dataset to assess the quality of depth maps. ETH3D benchmark~\cite{Schops2017Multi} consists of three scenarios corresponding to different tasks for (multi-view) stereo algorithms. It is more challenging for containing a diverse set of viewpoints and scene types. Here we only focus on high-resolution multi-view stereo dataset with images at a resolution of $6048\times4032$\footnote{In fact, we resize this imagery to no more than 3200 pixels for each dimension as \cite{Schonberger2016Pixelwise} does.}. Additionally, the high-resolution multi-view stereo dataset contains training datasets and test datasets. The training datasets provide not only ground truth point clouds but also ground truth depth maps, while the ground truth of the test datasets is withheld by the benchmark's web site.

All of our experiments are conducted on a machine with two Intel E5-2630 CPUs and two GTX Titan X GPUs. In the multi-hypothesis joint view selection scheme, $\{\tau_{0},\tau_{1},\alpha,\beta,n_{1},n_{2}\}=\{0.8,1.2,90,0.3,2,3\}$. In our geometric consistency guidance strategy and detail restorer, $\{k,\eta,\delta,\lambda,\xi\}=\{3,0.5,3,0.2,0.1\}$. Note that, we use only every other row and column in the window to speed up the computation of matching cost~\cite{Galliani2015Massively}.

\subsection{Depth Map Evaluation}

We evaluate our method's effectiveness on depth map estimation on Strecha dataset and ETH3D benchmark in this section. Following \cite{Hu2012Least}, we calculate the percentage of pixels with a absolute depth error less than $2cm$ and $10cm$ from the ground truth in Table~\ref{tab:S-depthmap}. To show the effectiveness of the structured region information, we replace the adaptive checkerboard sampling and multi-hypothesis joint view selection in ACMH with the diffusion-like propagation and top-k-winners-take-all view selection, denoted as DWTA. 

As can be seen, with the structured region information, ACMH performs better than DWTA and is also competitive with COLMAP~\cite{Schonberger2016Pixelwise} without geometric consistency. Furthermore, we see that ACMM surpasses ACMH by a noteworthy margin and almost achieves the best performance in this dataset. Note that, HerzJesu contains more low-textured areas and CMPMVS\cite{Jancosek2011Multi} performs a bit better than ACMM on it in the case of $2cm$. This is because CMPMVS is a global energy-based method that mainly focuses on weakly-supported surface. However, on the Fountain dataset that contains more details, ACMM is much better than CMPMVS in the case of $2cm$. We also note that, COLMAP is a representative algorithm among local methods. ACMM outperforms COLMAP in the case of $2cm$, although there is no significant difference in the case of $10cm$.

To reflect more challenges such as low-textured areas and thin structures in real-world scenes, we further compare our reconstructed depth maps with COLMAP\footnote{Note that, the depth maps of COLMAP are obtained with its default parameters and are unfiltered.} on the high-resolution multi-view training datasets of ETH3D benchmark in Table~\ref{tab:E-depthmap}. We see that ACMM clearly outperforms COLMAP in these challenging datasets, especially in some indoor datasets including poorly textured regions, such as kicker, office and pipes. Moreover, ACMH almost achieves the second-best performance. Figure~\ref{fig:depthmap} illustrates some examples of the depth maps estimated by COLMAP, DWTA, ACMH and ACMM. As can be seen, ACMH also performs better than DWTA and itself yields more robust results than COLMAP and DWTA in low-textured areas as it leverages the structured region information. Note that, although COLMAP outperforms DWTA and ACMH in some well-textured datasets such as court. and facade, it performs worse than DWTA and ACMH in some challenging datasets such as electro and office. This is because COLMAP cannot gain robust belief in challenging regions to infer pixelwise view selection. Combined with the multi-scale scheme, ACMM can further boost the estimation in these regions. Moreover, the details are also kept. 

\begin{table}[t]
	\caption{Point cloud evaluation on the high-resolution multi-view test datasets of ETH3D benchmark showing accuracy / completeness / $F_{1}$ score (in \%) at different evaluation thresholds (including $2cm$ and $10cm$). The related values are from \cite{Schops2017ETH3D}.}
	\centering
	\footnotesize
	\begin{tabular}{c|c|c|c}
		\hline
		& method & $2 cm$ & $10 cm$ \\
		\hline
		\multirow{6}{*}{indoor} & PMVS & 90.66 / 28.16 / 40.28 & 96.97 / 42.50 / 55.40 \\
		& Gipuma & 86.33 / 31.44 / 41.86 & 98.31 / 52.22 / 65.41 \\
		& LTVRE & {\bf 93.44} / 63.54 / 74.54 & {\bf 99.34} / 82.72 / 89.92 \\
		& COLMAP & 91.95 / 59.65 / 70.41 & 98.11 / 82.82 / 89.28 \\
		& ACMH & 91.14 / 64.81 / 73.93 & 98.76 / 82.61 / 89.42 \\
		& ACMM & 90.99 / {\bf 72.73} / {\bf 79.84} & 97.79 / {\bf 88.22} / {\bf 92.50} \\
		\hline
		\multirow{6}{*}{outdoor} & PMVS & 88.34 / 42.89 / 55.82 & 95.95 / 55.17 / 68.12 \\
		& Gipuma & 78.78 / 45.30 / 55.16 & 97.36 / 62.40 / 75.18 \\
		& LTVRE & 91.82 / 74.45 / 81.41 & 98.72 / 90.18 / 94.19 \\
		& COLMAP & {\bf 92.04} / 72.98 / 80.81 & 98.64 / 89.70 / 93.79 \\
		& ACMH & 83.96 / {\bf 80.03} / 81.77 & 97.51 / {\bf 90.57} / 93.87 \\
		& ACMM & 89.63 / 79.17 / {\bf 83.58} & {\bf 98.85} / 90.43 / {\bf 94.35} \\
		\hline
		\multirow{6}{*}{all} & PMVS & 90.08 / 31.84 / 44.16 & 96.71 / 45.67 / 58.58 \\
		& Gipuma & 84.44 / 34.91 / 45.18 & 98.07 / 54.77 / 67.86 \\
		& LTVRE & {\bf 93.04} / 66.27 / 76.25 & {\bf 99.18} / 84.59 / 90.99 \\
		& COLMAP & 91.97 / 62.98 / 73.01 & 98.25 / 84.54 / 90.40 \\
		& ACMH & 89.34 / 68.62 / 75.89 & 98.44 / 84.60 / 90.53 \\
		& ACMM & 90.65 / {\bf 74.34} / {\bf 80.78} & 98.05 / {\bf 88.77} / {\bf 92.96} \\
		\hline
	\end{tabular}
	\label{tab:pointcloud}
\end{table}     

\subsection{Point Cloud Evaluation}

In this section, fusion is imposed to get more consistent point clouds. We evaluate our point clouds on the high-resolution multi-view test datasets of ETH3D benchmark. 

Table~\ref{tab:pointcloud} lists the accuracy, completeness and  $F_{1}$ score of the point clouds estimated by PMVS~\cite{Furukawa2010Accurate}, Gipuma~\cite{Galliani2015Massively}, LTVRE~\cite{Kuhn2017TV}, COLMAP, ACMH and ACMM. All these methods show similar results in accuracy. In terms of $F_{1}$ score, ACMH is competitive with other methods for its good depth map estimation. And, ACMM outperforms other methods as it inherits the structured region property of ACMH and combines with the multi-scale scheme. Furthermore, ACMM obtains much higher completeness than other methods on indoor datasets that contain more low-textured areas. This is because ACMM perceives more credible information in these areas. As for outdoor datasets, ACMM achieves almost the same completeness as ACMH does. Figure~\ref{fig:pointcloud_test} illustrates some qualitative results achieved by these methods. It can be observed that, ACMM produces more complete point clouds especially in the challenging areas, \eg, red boxes shown in Figure~\ref{fig:pointcloud_test}.

\subsection{Runtime Performance}

We list the runtime of depth map generation for different methods that belong to the scope of PatchMatch Stereo in Table~\ref{tab:runtime}. All these methods are conducted on a single GPU through our same platform\footnote{Note that, all these methods use only every other row and column in the window to compute the matching cost.}. ACMH and Gipuma both converge after $6$ iterations while COLMAP adopts $10$ iterations. For ACMM, it needs $7$ iterations at the coarsest scale and $6$ iterations at other scales. As Table~\ref{tab:runtime} shows, ACMH is around $6\times$ faster than COLMAP. This is because the sequential propagation of COLMAP only updates the status of one row (column) of pixels at a time and its each iteration needs propagations in $4$ directions. Though ACMH and Gipuma both leverage the checkerboard propagation, ACMH is also faster than Gipuma. This is mainly because Gipuma employs a bisection refinement, which produces more unnecessary hypotheses to test. As for ACMM, it spends extra computational time on multi-scale geometric consistency scheme. However, ACMM takes no more than twice the runtime spent by ACMH as its geometric consistency at the coarser scales is conducted on downsampled images. Therefore, it is still about $3\times$ faster than COLMAP. 

\begin{table}[t]
	\caption{Runtime (in second) of depth map generation for different methods on Strecha dataset.}
	\centering
	\footnotesize
	\begin{tabular}{ccccccc}
		\hline
		dataset & \#images & Gipuma & COLMAP & ACMH & ACMM \\
		\hline
		Fountain & 11 & 235.58 & 1046.88 & \textbf{173.55} & 321.66 \\
		HerzJesu & 8 & 134.34 & 709.14 & {\bf 88.85} & 141.26 \\
		\hline
	\end{tabular}
	\label{tab:runtime}
\end{table}


\section{Conclusion}

In this work, we propose a novel multi-view stereo method for effective and efficient depth map estimation. Based on structured region information, we first present our basic MVS method with Adaptive Checkerboard sampling and Multi-Hypothesis joint view selection (ACMH). These strategies help to propagate good hypotheses as soon as possible and infer pixelwise view selection. Focusing on the depth estimation in low-textured areas, we further combine ACMH with our proposed multi-scale geometric consistency guidance scheme (ACMM). The multi-scale geometric consistency together with a detail restorer helps obtain more discrimination over low-textured areas while retaining fine details. In experiments, we demonstrate that our methods can obtain smooth and consistent depth map estimation together with complete dense 3D models while keeping a good efficiency, which shows promising applications of our methods.

\noindent\textbf{Acknowledgments.} This work was supported by the National Natural Science Foundation of China under Grant 61772213 and Grant 91748204.


{\small
\bibliographystyle{ieee_fullname}
\bibliography{egbib}

\begin{thebibliography}{10}\itemsep=-1pt

\bibitem{Bailer2012Scale}
Christian Bailer, Manuel Finckh, and Hendrik P.~A. Lensch.
\newblock Scale robust multi view stereo.
\newblock In {\em Proceedings of the European Conference on Computer Vision},
  pages 398--411, 2012.

\bibitem{Barnes2009PatchMatch}
Connelly Barnes, Eli Shechtman, Adam Finkelstein, and Dan~B Goldman.
\newblock Patchmatch: A randomized correspondence algorithm for structural
  image editing.
\newblock In {\em ACM SIGGRAPH}, pages 24:1--24:11, 2009.

\bibitem{Cremers2011Multiview}
D. Cremers and K. Kolev.
\newblock Multiview stereo and silhouette consistency via convex functionals
  over convex domains.
\newblock {\em IEEE Transactions on Pattern Analysis and Machine Intelligence},
  33(6):1161--1174, 2011.

\bibitem{Hernandez2004Silhouette}
Carlos~Hernández Esteban and Francis Schmitt.
\newblock Silhouette and stereo fusion for 3d object modeling.
\newblock {\em Computer Vision and Image Understanding}, 96(3):367 -- 392,
  2004.

\bibitem{Faugeras1998Variational}
O. Faugeras and R. Keriven.
\newblock Variational principles, surface evolution, pdes, level set methods,
  and the stereo problem.
\newblock {\em IEEE Transactions on Image Processing}, 7(3):336--344, 1998.

\bibitem{Furukawa2015MST}
Yasutaka Furukawa and Carlos Hern\'{a}ndez.
\newblock Multi-view stereo: A tutorial.
\newblock {\em Found. Trends. Comput. Graph. Vis.}, 9(1-2):1--148, 2015.

\bibitem{Furukawa2010Accurate}
Y. Furukawa and J. Ponce.
\newblock Accurate, dense, and robust multiview stereopsis.
\newblock {\em IEEE Transactions on Pattern Analysis and Machine Intelligence},
  32(8):1362--1376, 2010.

\bibitem{Galliani2015Massively}
S. Galliani, K. Lasinger, and K. Schindler.
\newblock Massively parallel multiview stereopsis by surface normal diffusion.
\newblock In {\em Proceedings of the IEEE International Conference on Computer
  Vision}, pages 873--881, 2015.

\bibitem{Goesele2007Multi}
M. Goesele, N. Snavely, B. Curless, H. Hoppe, and S.~M. Seitz.
\newblock Multi-view stereo for community photo collections.
\newblock In {\em Proceedings of the IEEE International Conference on Computer
  Vision}, pages 1--8, 2007.

\bibitem{Hartley2004Multiple}
Richard Hartley and Andrew Zisserman.
\newblock {\em Multiple View Geometry in Computer Vision}.
\newblock Cambridge University Press, 2 edition, 2004.

\bibitem{Hiep2009Towards}
V.~H. Hiep, R. Keriven, P. Labatut, and J. Pons.
\newblock Towards high-resolution large-scale multi-view stereo.
\newblock In {\em Proceedings of the IEEE Conference on Computer Vision and
  Pattern Recognition}, pages 1430--1437, 2009.

\bibitem{Hu2012Least}
X. Hu and P. Mordohai.
\newblock Least commitment, viewpoint-based, multi-view stereo.
\newblock In {\em International Conference on 3D Imaging, Modeling, Processing,
  Visualization Transmission}, pages 531--538, 2012.

\bibitem{Jancosek2011Multi}
M. Jancosek and T. Pajdla.
\newblock Multi-view reconstruction preserving weakly-supported surfaces.
\newblock In {\em Proceedings of the IEEE Conference on Computer Vision and
  Pattern Recognition}, pages 3121--3128, 2011.

\bibitem{Kanade1994Stereo}
T. Kanade and M. Okutomi.
\newblock A stereo matching algorithm with an adaptive window: theory and
  experiment.
\newblock {\em IEEE Transactions on Pattern Analysis and Machine Intelligence},
  16(9):920--932, 1994.

\bibitem{Kopf2007JBU}
Johannes Kopf, Michael~F. Cohen, Dani Lischinski, and Matt Uyttendaele.
\newblock Joint bilateral upsampling.
\newblock {\em ACM Trans. Graph.}, 26(3), 2007.

\bibitem{Kuhn2017TV}
Andreas Kuhn, Heiko Hirschm{\"u}ller, Daniel Scharstein, and Helmut Mayer.
\newblock A tv prior for high-quality scalable multi-view stereo
  reconstruction.
\newblock {\em International Journal of Computer Vision}, 124(1):2--17, 2017.

\bibitem{Lhuillier2005Quasi}
M. Lhuillier and L. Quan.
\newblock A quasi-dense approach to surface reconstruction from uncalibrated
  images.
\newblock {\em IEEE Transactions on Pattern Analysis and Machine Intelligence},
  27(3):418--433, 2005.

\bibitem{Bleyer2011PatchMatch}
Christoph~Rhemann Michael~Bleyer and Carsten Rother.
\newblock Patchmatch stereo - stereo matching with slanted support windows.
\newblock In {\em Proceedings of the British Machine Vision Conference}, pages
  14.1--14.11, 2011.

\bibitem{Schonberger2016Pixelwise}
Johannes~L. Sch{\"o}nberger, Enliang Zheng, Jan-Michael Frahm, and Marc
  Pollefeys.
\newblock Pixelwise view selection for unstructured multi-view stereo.
\newblock In {\em Proceedings of the European Conference on Computer Vision},
  pages 501--518, 2016.

\bibitem{Schops2017ETH3D}
Thomas Sch{\"o}ps, Johannes~L. Sch{\"o}nberger, Silvano Galliani, Torsten
  Sattler, Konrad Schindler, Marc Pollefeys, and Andreas Geiger.
\newblock Eth3d benchmark.
\newblock \url{https://www.eth3d.net}.

\bibitem{Schops2017Multi}
T. Sch{\"o}ps, J.~L. Sch{\"o}nberger, S. Galliani, T. Sattler, K. Schindler, M.
  Pollefeys, and A. Geiger.
\newblock A multi-view stereo benchmark with high-resolution images and
  multi-camera videos.
\newblock In {\em Proceedings of the IEEE Conference on Computer Vision and
  Pattern Recognition}, pages 2538--2547, 2017.

\bibitem{Seitz2006Comparison}
S.~M. Seitz, B. Curless, J. Diebel, D. Scharstein, and R. Szeliski.
\newblock A comparison and evaluation of multi-view stereo reconstruction
  algorithms.
\newblock In {\em Proceedings of the IEEE Conference on Computer Vision and
  Pattern Recognition}, volume~1, pages 519--528, 2006.

\bibitem{Shan2013Visual}
Q. Shan, R. Adams, B. Curless, Y. Furukawa, and S.~M. Seitz.
\newblock The visual turing test for scene reconstruction.
\newblock In {\em International Conference on 3D Vision}, pages 25--32, 2013.

\bibitem{Shan2014Occluding}
Q. Shan, B. Curless, Y. Furukawa, C. Hernandez, and S.~M. Seitz.
\newblock Occluding contours for multi-view stereo.
\newblock In {\em Proceedings of the IEEE Conference on Computer Vision and
  Pattern Recognition}, pages 4002--4009, 2014.

\bibitem{Shen2013Accurate}
S. Shen.
\newblock Accurate multiple view 3d reconstruction using patch-based stereo for
  large-scale scenes.
\newblock {\em IEEE Transactions on Image Processing}, 22(5):1901--1914, 2013.

\bibitem{Sinha2007Multi}
S.~N. Sinha, P. Mordohai, and M. Pollefeys.
\newblock Multi-view stereo via graph cuts on the dual of an adaptive
  tetrahedral mesh.
\newblock In {\em Proceedings of the IEEE International Conference on Computer
  Vision}, pages 1--8, 2007.

\bibitem{Strecha2008On}
C. Strecha, W. von Hansen, L.~Van Gool, P. Fua, and U. Thoennessen.
\newblock On benchmarking camera calibration and multi-view stereo for high
  resolution imagery.
\newblock In {\em Proceedings of the IEEE Conference on Computer Vision and
  Pattern Recognition}, pages 1--8, 2008.

\bibitem{Tylecek2010Refinement}
Radim Tylecek and R Sara.
\newblock Refinement of surface mesh for accurate multiview reconstruction.
\newblock {\em International Journal of Virtual Reality}, 9(1):45--54, 2010.

\bibitem{Vogiatzis2007Multiview}
G. Vogiatzis, C.~Hernández Esteban, P.~H.~S. Torr, and R. Cipolla.
\newblock Multiview stereo via volumetric graph-cuts and occlusion robust
  photo-consistency.
\newblock {\em IEEE Transactions on Pattern Analysis and Machine Intelligence},
  29(12):2241--2246, 2007.

\bibitem{Wei2014Multi}
Jian Wei, Benjamin Resch, and Hendrik Lensch.
\newblock Multi-view depth map estimation with cross-view consistency.
\newblock In {\em Proceedings of the British Machine Vision Conference}, 2014.

\bibitem{Yoon2005Locally}
Kuk-Jin Yoon and In-So Kweon.
\newblock Locally adaptive support-weight approach for visual correspondence
  search.
\newblock In {\em Proceedings of the IEEE Conference on Computer Vision and
  Pattern Recognition}, volume~2, pages 924--931 vol. 2, 2005.

\bibitem{Zach2008Fast}
Christopher Zach.
\newblock Fast and high quality fusion of depth maps.
\newblock In {\em International Conference on 3D Imaging, Modeling, Processing,
  Visualization Transmission}, 2008.

\bibitem{Zaharescu2011Topology}
A. Zaharescu, E. Boyer, and R. Horaud.
\newblock Topology-adaptive mesh deformation for surface evolution, morphing,
  and multiview reconstruction.
\newblock {\em IEEE Transactions on Pattern Analysis and Machine Intelligence},
  33(4):823--837, 2011.

\bibitem{Zhang2008Recovering}
Guofeng Zhang, Jiaya Jia, Tien-Tsin Wong, and Hujun Bao.
\newblock Recovering consistent video depth maps via bundle optimization.
\newblock In {\em Proceedings of the IEEE Conference on Computer Vision and
  Pattern Recognition}, pages 1--8, 2008.

\bibitem{Zhang2009Cross}
K. Zhang, J. Lu, and G. Lafruit.
\newblock Cross-based local stereo matching using orthogonal integral images.
\newblock {\em IEEE Transactions on Circuits and Systems for Video Technology},
  19(7):1073--1079, 2009.

\bibitem{Zheng2014PatchMatch}
E. Zheng, E. Dunn, V. Jojic, and J.~M. Frahm.
\newblock Patchmatch based joint view selection and depthmap estimation.
\newblock In {\em Proceedings of the IEEE Conference on Computer Vision and
  Pattern Recognition}, pages 1510--1517, 2014.

\end{thebibliography}
}

\end{document}